\documentclass[11pt]{article}

\usepackage[final]{acl}

\usepackage{times}
\usepackage{latexsym}

\usepackage[T1]{fontenc}

\usepackage[utf8]{inputenc}

\usepackage{microtype}

\usepackage{inconsolata}

\usepackage{graphicx}

\usepackage{booktabs}
\usepackage{amssymb}
\usepackage{tcolorbox}
\tcbuselibrary{breakable}

\usepackage{amsmath}
\usepackage{subcaption}

\usepackage{dialogue}
\usepackage{longtable}
\usepackage{colortbl}
\usepackage{float}
\usepackage{adjustbox}
\usepackage{listings}
\usepackage{pgfplots}
\usepackage{placeins}
\usetikzlibrary{patterns}
\pgfplotsset{compat=1.15}

\definecolor{babyblue}{rgb}{0.54, 0.81, 0.94}
\definecolor{babyred}{RGB}{252, 165, 165}
\usepackage{enumitem}
\usepackage[normalem]{ulem}

\title{Revising Context, Shifting Simulated Stance: Auditing LLM-Based Stance Simulation in Online Discussions}

\author{Xinnong Zhang\textsuperscript{\rm 1,4}\thanks{These authors contribute equally to this work.}, 
        Wanting Shan\textsuperscript{\rm 2}\footnotemark[1], 
        Hanjia Lyu\textsuperscript{\rm 3}\thanks{Project lead.}, 
        Zhongyu Wei\textsuperscript{\rm 1,4},
        Jiebo Luo\textsuperscript{\rm 2} \\
        \textsuperscript{\rm 1}{Fudan University}, \textsuperscript{\rm 2}{University of Rochester}, \\
        \textsuperscript{\rm 3}{Singapore Management University},
        \textsuperscript{\rm 4}{Shanghai Innovation Institute}
        \\
        \texttt{xnzhang23@m.fudan.edu.cn}, \texttt{wshan2@u.rochester.edu}, \\
         \texttt{hjlyu@smu.edu.sg}, \texttt{zywei@fudan.edu.cn}, \texttt{jluo@cs.rochester.edu}
        }

\begin{document}
\maketitle
\begin{abstract}
Large language models are increasingly used to simulate social media users and infer how individuals may respond to online discussions. However, it remains unclear whether these simulations reflect precise user-specific beliefs or whether they are highly sensitive to semantically independent changes in conversational contexts. In this work, we study counterfactual context revision as a framework for auditing LLM-based stance simulation. Given an original online conversation, we first infer a target user’s stance toward a specific topic. We then apply controlled revision strategies to the conversational context and simulate the user’s stance again under the revised context. We compare text-only revision strategies with a multimodal one that incorporates meme-based context and evaluate two main effectiveness metrics, \textit{i.e.,} average directional stance shift and stance transition rate. The results reveal effective and robust stance transitions in both text-only and multimodal strategies across different polarization-preference mechanisms. Our study contributes an evaluation framework for understanding the context sensitivity of LLM-based stance simulation (\url{https://github.com/STARResearchLab/StanceShift}). More broadly, it highlights both the promise and risk of using LLMs as proxies for social media users in social simulation.
\end{abstract}

\begin{figure}[t]
    \centering
    \includegraphics[width=0.9\columnwidth]{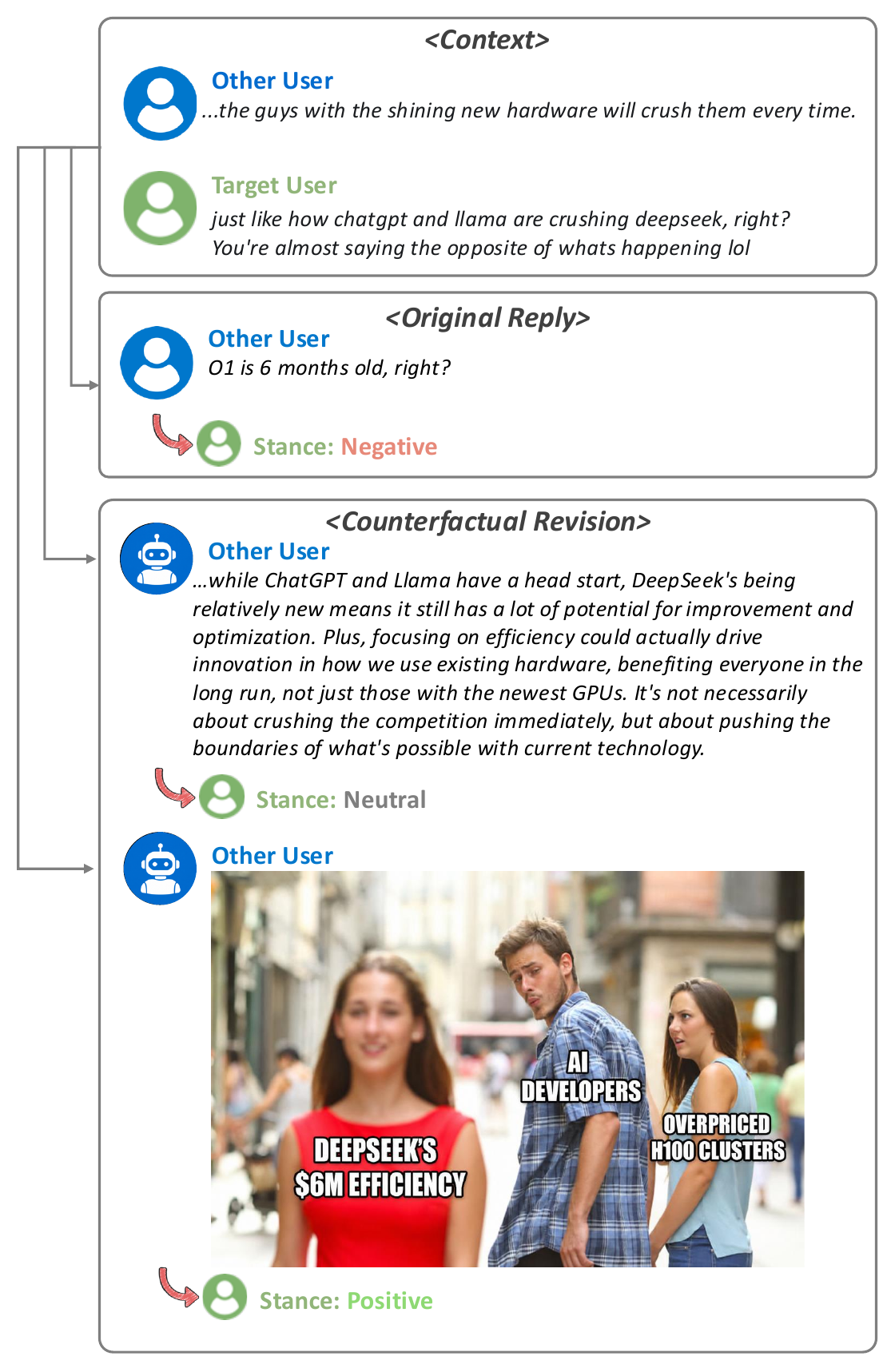}
    \caption{An example showing the LLM-based simulated stance shift when applying different counterfactual revision strategies.}
    \label{fig:head-pic}
\end{figure}

\section{Introduction}

Large language models have rapidly become a promising tool for studying online social behavior. In computational social science (CSS), researchers have begun using LLMs to simulate social media users, infer public opinions, predict responses in online discussions, and model how individuals may react to different informational environments~\citep{argyle2023out, park2023generative, chuang2024simulating, qiu2025can}. These applications are appealing because LLMs can process rich conversational context, generate human-like responses, and provide scalable approximations of social interaction~\citep{aher2023using, gao2024large}. As a result, LLM-based simulation is increasingly viewed as a potential complement to traditional surveys, annotation studies, and observational analyses of online communities~\citep{ziems2024can, gilardi2023chatgpt}.

However, the use of LLMs for social simulation also raises a fundamental question: \emph{what exactly is being simulated?} When an LLM predicts a user’s stance in an online discussion, the output may reflect the target user’s expressed preferences, the conversational evidence available in the thread, the model’s prior assumptions about the topic, or superficial cues introduced by the prompt. This issue becomes especially important in stance simulation, where the task is not merely to classify sentiment, but to infer whether a user supports, opposes, or remains neutral toward a specific target~\citep{kucuk2020stance, zhang2024llm, zhao2024ezstance}. If simulated stances are highly susceptible to small revisions in the surrounding context, this suggests that LLM-based simulations may reflect context-sensitive model behaviors more than stable approximations of human opinion~\citep{santurkar2023whose, rottger2024political, sclar2024quantifying}.

In this paper, we study this issue through the lens of counterfactual context revision. Our central idea is simple: given an original online conversation involving a target user, we first use an LLM to infer the user’s stance from the original context. We then revise the conversational context using controlled strategies while keeping the target user and stance target topic fixed. Finally, we ask the LLM to simulate the user’s stance again under the revised context. This setup allows us to move beyond the question of whether LLMs can infer stance from a single static conversation. Instead, we ask \emph{how simulated stance changes when the textual or multimodal context surrounding the user is altered}. We emphasize that our goal is not to assess whether simulated stance shifts reflect real-world opinion change, which would require comparison with observed human behavior. Rather, we audit whether LLM-based stance simulations are robust or overly sensitive to controlled revisions of the conversational context. Accordingly, the relevant comparison is between stances produced by the same simulator across different contexts.

We examine both text-only and multimodal context revision strategies. 
Multimodal revisions are especially important for social media, where opinions are often shaped not only by text, but also by images, memes, screenshots, reaction images, and other visual signals. A meme or image can introduce humor, emotional framing, group identity, or implicit evaluation that is difficult to capture through text alone. By comparing multimodal revisions against their text-only variants, we discover that visual context provides additional influence on simulated stance beyond its textual description.

\begin{figure*}
    \centering
    \includegraphics[width=\linewidth]{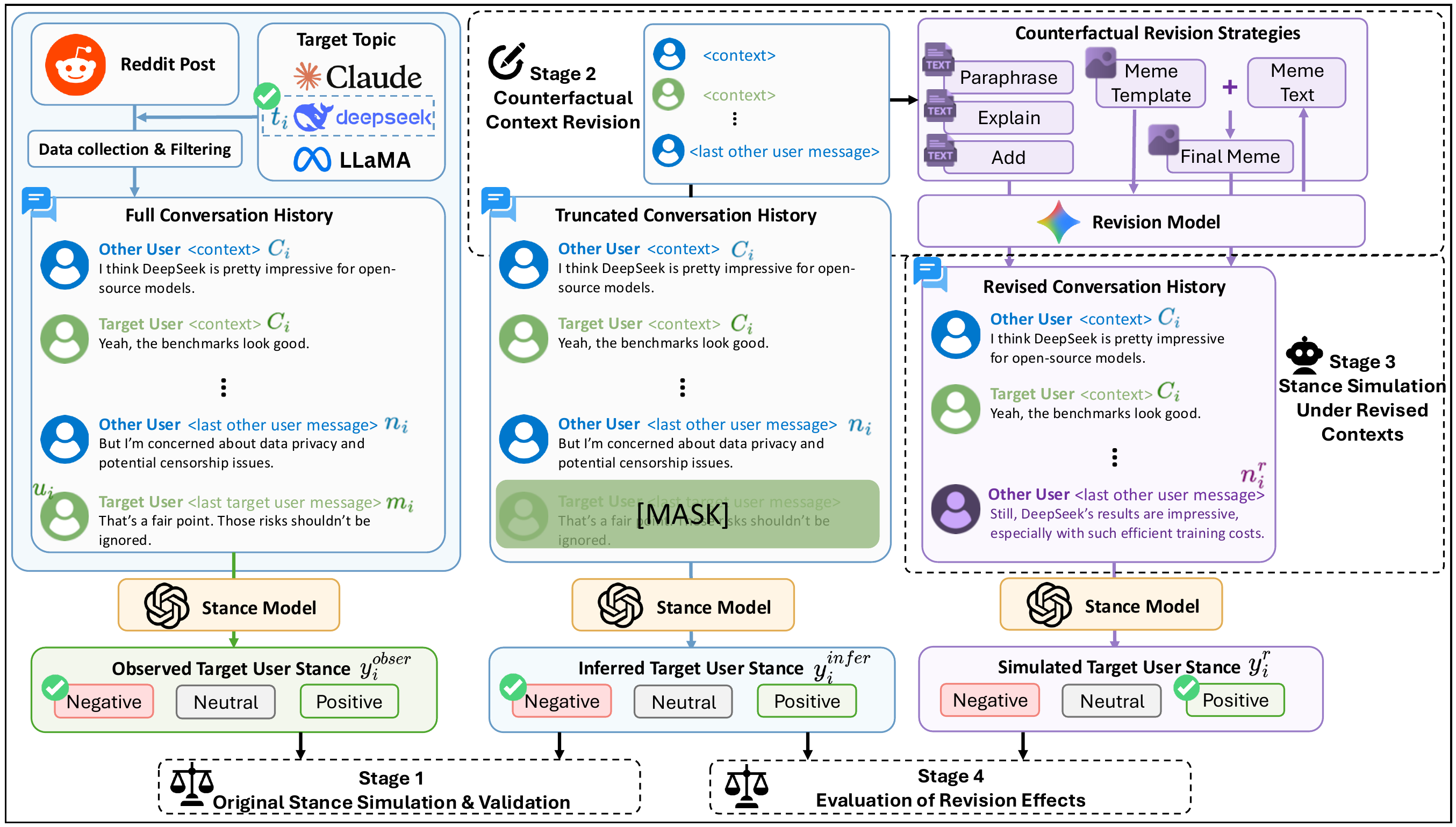}
    \caption{Overview of the study design and experiment setup.}
    \label{fig:framework}
\end{figure*}

Through extensive experiments and analysis, we discover that LLM-based stance simulation remains overall robust across three stance topics and different simulation models. We find that adding extra information to the simulation context can help change the simulated stance, which can be achieved through either text-only \textit{add} strategy or multimodal \textit{meme} strategy. Additionally, we find different revision mechanisms in text-only and multimodal strategies, with a significant depolarized trend in \textit{add} and a strong polarized trend in \textit{meme}.

This paper makes the following contributions. First, we introduce counterfactual context revision as a framework for auditing LLM-based stance simulation in online discussions. Second, we systematically compare text-only and multimodal revision strategies for shifting simulated user stance. 
Third, we provide a mechanism-level analysis of why different revision strategies influence simulated stance.
Together, these contributions provide a more nuanced understanding of the opportunities and limitations of LLM-based social simulation.

\section{Study Design and Experimental Setup}
\label{sec:study_design}

We design our study to evaluate how LLM-based stance simulations respond to controlled revisions of the conversational context surrounding social media users. Rather than treating stance simulation as a one-step prediction task, we formulate it as a counterfactual context revision problem. As shown in Figure~\ref{fig:framework}, given an original online discussion, we first infer a target user's stance toward a specific discussion target stance topic. We then revise the surrounding conversation using different revision strategies and simulate the same user's stance again under each revised context. This design allows us to audit not only simulated stance changes but also the most effective types of revision strategies and potential mechanisms that explain the shift.

\subsection{Data Collection and Preprocessing}

To study counterfactual context revision in a realistic and socially relevant setting, we focus on online discussions about emerging AI technologies. 
More specifically, we collect Reddit discussions related to three major LLM families: DeepSeek, Claude, and Llama. Details of data collection and preprocessing are described in Appendix~\ref{appendix_sec:data}.
The final dataset contains 1,821 conversation instances, including 787 discussions related to DeepSeek, 538 related to Claude, and 496 related to Llama. These conversations are collected from 97 unique subreddits and 416 unique Reddit posts, involving 851 unique target users and 2,945 unique authors participating in the conversation threads.

\subsection{Task Formulation}\label{subsec:task-formation}

For each conversation instance $i$, let $m_i$ denote the last target user message and $n_i$ denote the last other user message. Next, let $C_i$ denote the rest of the original conversational context apart from $(m_i, n_i)$, $u_i$ denote the target user, and $t_i$ denote the stance target topic, such as DeepSeek, Claude, or Llama. The goal of the stance model (\textit{i.e.,} simulator) is to infer the target user's stance toward $t_i$ given the context $C_i$ and $n_i$. We consider a three-way stance label space: positive, neutral, and negative, which captures whether a user expresses support for, neutrality toward, or opposition to a given topic.

Given the full conversation history $(C_i, n_i, m_i)$, the stance of the target user is labeled as the observed stance:
\begin{equation}
    y_i^{obser} = f_{\theta}(u_i, t_i \mid C_i, n_i, m_i),
\end{equation}
where $f_{\theta}$ is an LLM-based stance simulator. The original conversation history without $m_i$ produces an inferred stance prediction:
\begin{equation}
    y_i^{infer} = f_{\theta}(u_i, t_i \mid C_i, n_i),
\end{equation}
We then apply a revision strategy $r$ to the context to produce a revised last other user message:
\begin{equation}
    n_i^{r} = g_\theta^r(u_i, t_i \mid C_i, n_i),
\end{equation}
where $g_\theta^r$ is a context revision LLM model guided by a specific strategy. The stance simulator is then applied again:
\begin{equation}
    y_i^{r} = f_{\theta}(u_i, t_i \mid C_i, n_i^r).
\end{equation}

We focus mainly on the change between $y_i^{infer}$ and $y_i^{r}$, which indicates how the simulated stance of the same target user shifts when the conversational context is revised. The prompts for inferring the stance are described in Appendix~\ref{app:stance-prompt}.

\subsection{Stage 1: Original Stance Simulation and Validation}

The first stage evaluates whether LLMs can reasonably infer user stance from the original Reddit conversations. According to the task formation in Section~\ref{subsec:task-formation} and Figure~\ref{fig:framework}, for each instance, given the original conversational context $C_i$, last other user message $n_i$, last target user message $m_i$, the target user $u_i$, and the stance target $t_i$. The model is asked to annotate whether the target user is supportive, neutral, or opposing toward the target. To reduce ambiguity, the prompt explicitly distinguishes target-specific stance from general sentiment. We employ the observed stance $y_i^{obser}$ as the basis for validation.

After masking the $m_i$, the stance model infers the stance $y_i^{infer}$ of the target user. We then compare the inferred stance $y^{infer}$ with the observed stance $y^{obser}$ to assess the baseline validity of the simulation. This stage is important because counterfactual revision analysis is only meaningful if the stance model can first produce reasonable stance estimates under the original context.

We report standard stance classification metrics such as accuracy, macro F1, and weighted F1. Macro F1 is used as a primary evaluation metric given that stance labels may be imbalanced. We also analyze performance separately across the three targets, DeepSeek, Claude, and Llama. This helps determine whether the simulator performs consistently or whether it is more reliable for some model communities than others.

\subsection{Stage 2: Counterfactual Context Revision}

The second stage applies controlled revision strategies to the original conversational context. The purpose of this stage is not to generate arbitrary persuasive content, but to create counterfactual versions of the same discussion that differ along interpretable contextual dimensions~\cite{kaushik2019learning}. Each revision strategy modifies the last other user message $n_i$ while preserving the main topic, conversational coherence, and the identity of the stance target.

We consider both text-only and multimodal revision strategies. Text-only strategies revise the conversation through language-based changes, while the multimodal strategy generates a meme to replace the original $n_i$.
We also include controlled revisions. A minimal \textit{paraphrase} strategy rewrites the context without intentionally changing its stance-relevant content. 
For the multimodal setting, we use text-only counterparts as the controlled strategies, allowing us to test whether the visual modality itself contributes additional influence beyond a textual description of the same information.

To verify that the generated revisions are controlled, meaning-preserving interventions rather than uncontrolled LLM generations, we conduct a human validation study. Three annotators evaluate 540 revision instances for strategy adherence, contextual coherence, and target preservation, yielding high inter-annotator agreement. The protocol and full results are provided in Appendix~\ref{app:human-validation}.

\subsection{Stage 3: Stance Simulation Under Revised Contexts}

The third stage applies the same stance simulator to each revised context. For each original conversation, we obtain one prediction $y^{infer}$ under the original context and one prediction $y^{r}$ under each revised context. The simulator receives the same task instruction, target user identifier, and stance target across all conditions. The only difference is the surrounding conversational context.

For each prediction, the simulator outputs a stance label and a short explanation. The explanation is \emph{not} used as direct evidence of model reasoning, but it helps support qualitative analysis of how the model interprets the revised context.
We repeat the simulation across different revision models and stance models to evaluate whether revision effects are model-specific or robust across different LLM families. Stage 3 produces a matrix of stance predictions for each conversation:
\begin{equation}
    Y_i = \{y_i^{infer}, y_i^{r_1}, y_i^{r_2}, ..., y_i^{r_k}\},
\end{equation}
where each $r_k$ corresponds to a different revision strategy. This structure allows us to compare strategies within the same original conversation, target user, and stance target.

\subsection{Stage 4: Evaluation of Revision Effects}
First, we measure effectiveness using two complementary metrics. The first metric is the average directional stance shift. We map stance labels onto an ordinal scale, negative is $-1$, neutral is $0$, and positive is $1$~\citep{mohammad2016semeval}. For each revised context, we compute:
\begin{equation}
    \Delta_i^r = score(y_i^r) - score(y_i^{infer}).
\end{equation}
A positive value indicates that the revised context moves the simulated stance in a more supportive direction, while a negative value indicates movement in a more opposing direction. We report the average $\Delta_i^r$ for each revision strategy as a summary measure of its overall directional effect.

The second metric is the stance change rate, which captures specific transition patterns between the original and revised simulations. We identify two types of changes: a supportive stance change, including negative to neutral/positive and neutral to positive, and an opposing stance change, including neutral to negative, indicating the backfire effect~\citep{nyhan2010corrections}. Formally, the stance change rate $\mathrm{R}$ over a revision strategy $r$ can be described as:
\begin{equation}
\begin{aligned}
R_{a \rightarrow b}^r
&=
\frac{
\sum_{i=1}^{N}
\mathbb{I}\!\left(
y_i^{infer} = a
\land
y_i^{r} = b
\right)
}{
\sum_{i=1}^{N}
\mathbb{I}\!\left(
y_i^{infer} \in \{\mathrm{neg,neu}\}
\right)
}, \\
&\text{where }a,b \in \{\mathrm{neg, neu, pos}\}
\end{aligned}
\end{equation}
These two metrics evaluate both the overall direction of stance movement and the concrete forms of stance change produced by each revision strategy.

Second, we evaluate robustness. We test whether revision effects hold across the three stance targets, across different subreddits, across discussion topics, and across simulator models. We also examine whether effects are stable under prompt variations and decoding settings. A robust revision strategy should produce consistent patterns across these conditions, rather than only working for one model, one prompt template, or one target community.

\subsection{Meme-Based Multimodal Revision}

In addition to text-only revision strategies, we include a meme-based multimodal revision strategy. This condition is motivated by the fact that online discussions often use memes to express stances indirectly through humor, affect, cultural references, and visual framing. Unlike purely textual revisions, a meme can reinforce or reinterpret the surrounding conversation by combining image content with short textual cues. This makes memes a useful case for studying whether multimodal social media context changes LLM-based stance simulation beyond text-only conversational edits.

For each selected conversation, the meme-based revision employs a stance-relevant meme to replace the last other user message $n_i$ while keeping the rest unchanged. The meme is guided to be coherent with the discussion and to express a contextual frame related to the target. This allows us to test whether meme-based framing changes the simulated stance of the target user under a comparable context setting.

We evaluate the meme-based strategy against text-only revision strategies using the same effectiveness and robustness metrics described above. In particular, we examine whether meme-based revision produces larger or more consistent directional stance shifts, whether it reduces opposition or increases support, and whether it introduces higher risks of backfire. This comparison allows us to assess whether meme-based multimodal context provides additional value in LLM-based stance simulation, while avoiding a broader claim about multimodality in general.

\section{Revision Strategies}
All the prompts can be found in Appendix~\ref{app:revision-prompt}. Appendix~\ref{appendix_sec:implementation} describes the implementation details.
\paragraph{Paraphrase}
 aims to reformulate the last message using similar contextual language without intentionally introducing new arguments and interpretations. 
This strategy is intended to simulate the variations that exist in language while the underlying argumentation content remains the same, in order to investigate whether LLM stance inference is sensitive to wording differences. 

\paragraph{Explain}
 is designed to acknowledge any misunderstanding or ambiguity present within the entire conversation. It does not involve altering the message but rather expanding on the reasoning and clarifications included within the message.
The purpose behind this approach is to understand whether better explanations, acknowledgment of concerns, and more constructive reasoning influence how stance models understand the position held by the target user. 

\paragraph{Add}
attempts to review the last message $n_i$ within the conversation and provide extra arguments or perspectives to address the problems expressed by the target user. In contrast with the \textit{paraphrase} and \textit{explain} strategies, which focus only on clarifications, the \textit{add} strategy aims to extend the conversational content, adding arguments that may encourage a more positive interpretation of the discussed model while retaining its factual consistency with the initial conversation. 
The goal of applying this strategy is to investigate whether the addition of extra arguments or supportive framing helps to persuade stance models, even when the target user's original statements remain unchanged.

\paragraph{Meme}
is a multimodal revision strategy that assesses the simulated target user's stance through two steps: meme text generation and meme generation. In the meme text generation step, the conversation history is provided to the revision LLM along with a meme template. The revision LLM is asked to generate appropriate meme text tailored to the conversation context. In the meme generation step, the meme text and meme template are fed into a multimodal language model to generate the final meme. Altogether, 5 meme templates are collected from ImgFlip\footnote{\url{https://imgflip.com/memetemplates}} and used in the main experiments. All meme templates are shown in Appendix~\ref{app:meme-temp}.

\section{Results}
\label{sec:results}

We organize the results around three questions. First, we evaluate whether LLMs can infer user stance from the original Reddit conversations before any context revision is applied. Second, we compare how different text-only revision strategies change simulated stance. Third, we examine whether the meme-based multimodal revision produces additional effects beyond text-only revisions. 

\begin{table}[t]
\centering
\resizebox{\columnwidth}{!}{
\begin{tabular}{lccc}
\toprule
\textbf{Target} & \textbf{Accuracy} & \textbf{Macro F1} & \textbf{Weighted F1} \\
\midrule
DeepSeek & 77.38 & 77.93 & 77.39 \\
Claude   & 78.58 & 78.25 & 78.87 \\
Llama    & 77.02 & 76.05 & 76.81 \\ \midrule
Overall  & 77.64 & 78.10 & 77.81 \\
\bottomrule
\end{tabular}}
\caption{Evaluation of original stance simulation. Macro F1 is the primary metric due to class imbalance.}
\label{tab:original_stance_performance}
\end{table}

\begin{table*}[t]
\centering
\small
\setlength{\tabcolsep}{4pt}
\resizebox{0.9\linewidth}{!}{
\begin{tabular}{@{}lccccc@{}}
\toprule
{\textbf{Target Topic}} & \textbf{Inferred} & \textbf{paraphrase \(\Delta\)} & \textbf{explain \(\Delta\)} & \textbf{add \(\Delta\)} & \textbf{meme \(\Delta\)} \\ \midrule
Claude & 0.234 & +0.002 (+0.8\%) & +0.056 (+23.8\%) & +0.113 (+48.4\%) & +0.141 (+60.3\%) \\
Deepseek & 0.170 & -0.011 (-6.7\%) & -0.008 (-4.5\%) & +0.029 (+17.2\%) & +0.076 (+44.4\%) \\
Llama & 0.133 & -0.010 (-7.6\%) & +0.067 (+50.0\%) & +0.125 (+93.9\%) & +0.051 (+38.6\%) \\ \midrule
Average &  0.179 & -0.007 (-4.0\%) & +0.031 (+17.5\%) & +0.080 (+44.8\%) & +0.088 (+49.3\%) \\
\bottomrule
\end{tabular}}
\caption{The average directional stance shift of different strategies on all target topics and stance models, stanced by GPT-5.2. The full table is shown in Table~\ref{tab:app-main-delta}.}
\label{tab:main-delta}
\end{table*}

\begin{figure*}[t]
    \centering
    \includegraphics[width=\linewidth]{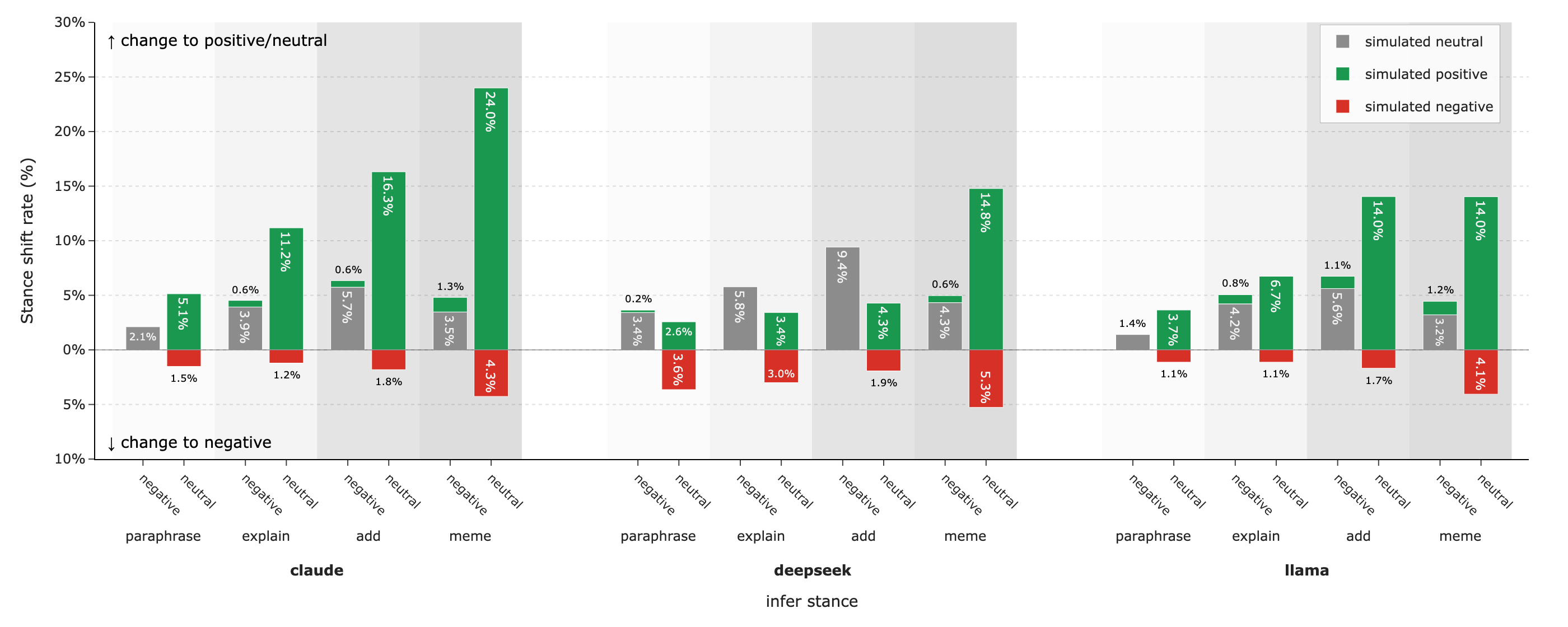}
    \caption{An illustration of the stance transition rates of different strategies on all target topics stanced by GPT-5.2. Full illustration is shown in Figure~\ref{fig:app-3-models}.}
    \label{fig:main-transition-rate}
\end{figure*}

\subsection{Original Conversations Provide a Baseline for Stance Simulation}
\label{sec:results_original}

We first evaluate stance simulation under the original conversational context. This step is necessary because counterfactual revision analysis is meaningful only if the simulator can produce reasonable stance estimates before any revision is introduced. For each conversation, the simulator is given the original context, the target user, and the discussion target, and is asked to infer whether they are opposing, neutral, or supportive toward the target.

As shown in Table~\ref{tab:original_stance_performance}, the simulator achieves a macro F1 of 78.10 and an accuracy of 77.64 between $y^{obser}$ and $y^{infer}$. Performance remains stable across three target topics, with the highest macro F1 on Claude (78.25) and the lowest on Llama (76.05). 
The relatively small gap suggests that the original stance simulation is not strongly target-dependent and $y^{infer}$ can provide a consistent approximation of the $y^{obser}$. This supports the use of $y_i^{{infer}}$ as a reasonable proxy for $y_i^{{obser}}$ in the subsequent counterfactual revision analysis.

Overall, the original stance simulation results provide a baseline for the counterfactual revision analysis. The simulator is able to recover a substantial portion of observed stance patterns, but its errors also reveal ambiguity in technology-related discussions. We therefore interpret \textit{the revision results as changes in LLM simulated stance, rather than as direct evidence of real user opinion change}.

\subsection{Context Revisions Produce Distinct Patterns of Simulated Stance Shift}
\label{sec:results_revision_overall}

We next evaluate how simulated stance changes after applying each revision strategy. For each strategy, we compare the stance predicted under the revised context with the stance inferred under the original context. We report two main effectiveness metrics: average directional stance shift and stance transition rate.

As shown in Table~\ref{tab:main-delta}, among the text-only strategies, \textit{add} produces the largest average positive shift, with an average $\Delta$ of +44.8\%. \textit{Explain} produces a smaller but still positive shift, while \textit{paraphrase} has a limited effect. These differences suggest that not all context revisions are equally influential for LLM-based stance simulation. Strategies that modify the evidential or interpretive structure of the conversation appear to have stronger effects than strategies that only change surface wording.

On the other hand, the meme-based revision produces an average directional shift of +49.3\%, compared with +44.8\% for the \textit{add} strategy and -4\% for the paraphrase control. This suggests that meme-based context can influence LLM-based stance simulation, although the magnitude of the effect depends on the target and conversation type.

The stance transition rates provide a more interpretable view of these effects. As shown in Figure~\ref{fig:main-transition-rate}, \textit{add} is especially effective at moving originally opposing predictions to neutral or supportive predictions, with an average negative reduction rate of 4.7\%. In contrast, \textit{meme} is more effective at moving originally neutral predictions to positive predictions, with an average neutral to supportive rate of 17.6\%. This distinction is important because the two strategies appear to operate through different forms of simulated stance change: one primarily softens opposition, while the other activates support among initially neutral cases.

We also observe that stronger revision effects are not always preferable. \textit{Meme} produces larger stance shifts but also higher backfire rates, meaning that a portion of originally positive or neutral predictions become more opposing after revision, suggesting that effectiveness should be interpreted together with robustness, rather than only by the magnitude of stance shift.

\begin{table*}[t]
    \centering
    \adjustbox{width=\linewidth}{
    \begin{tabular}{llll}
    \toprule[1.1pt]
     \textbf{Method}  & \textbf{Revision Input} & \textbf{Inference Input} & \textbf{Purpose} \\\midrule
     r\_meme &  Meme template & Meme image + text &  Full method\\
     r\_white\_meme & Meme template & White background + text & Test visual information during inference\\
     r\_humor & Humor instruction & Meme image + text & Replace template with humor guidance\\
     r\_caption\_cut & Meme caption & Meme image + text & Replace a template with visual description \\
     r\_caption & Meme caption + usage knowledge & Meme image + text & Replace template with textualized meme knowledge\\
    \bottomrule[1.1pt]
    \end{tabular}
    }
    \caption{Ablation variants used to isolate the role of meme templates during revision generation and stance inference.
    }
    \label{tab:meme_variant}
\end{table*}

\begin{table}[t]
\small
\centering
\adjustbox{width=\linewidth}{
\begin{tabular}{@{}lcccc@{}}
\toprule
\textbf{Strategy} & \textbf{Claude} & \textbf{DeepSeek} & \textbf{Llama} & \textbf{Average} \\ \midrule
r\_meme & 21.9\% & 26.7\% & 20.6\% & 23.1\% \\
r\_white\_meme & 18.1\% & 22.2\% & 24.5\% & 21.6\% \\ \midrule
r\_humor & 12.9\% & 18.5\% & 25.8\% & 19.1\% \\
r\_caption\_cut & 18.1\% & 21.5\% & 20.6\% & 20.1\% \\
r\_caption & 16.8\% & 22.2\% & 21.9\% & 20.3\% \\
 \bottomrule
\end{tabular}
}
\caption{Results of combined stance transition rates of neutral and positive transition across different meme-based strategy variants.}
\label{tab:meme-ablation}
\end{table}

\newcommand{\tonepanelw}{1.9cm}
\newcommand{\tonepanelh}{2.6cm}
\begin{figure}[t]
    \centering
    \captionsetup[subfigure]{font=scriptsize,skip=1pt}
    \begin{subfigure}[b]{0.37\linewidth}
    \centering
    \begin{tikzpicture}
    \begin{axis}[
    scale only axis,
    width=\tonepanelw,
    height=\tonepanelh,
    ytick style={draw=none},
    xtick style={draw=none},
    ybar,
    bar width=3.5pt,
    xlabel={Tone Score},
    label style={font=\scriptsize},
    ylabel={Share of Comments (\%)},
    xmin=0, xmax=100,
    ymin=0, ymax=50,
    xtick={0,50,100},
    xticklabels={{\hspace{5pt}0},{50},{100\hspace{11pt}}},
    ytick={0,25,50},
    xticklabel style={
        font=\scriptsize
    },
    yticklabel style={
        font=\scriptsize
    },
    scaled y ticks=false,
    grid=major,
    legend style={at={(0.5,1.35)}, anchor=north, legend columns=-1, draw=none, nodes={inner sep=10pt}},
    legend image post style={
        xscale=0.5,
        yscale=1.5
    },
]
\addplot+[ybar, bar shift=0pt, fill=gray!35, draw=black, area legend] coordinates {
    (5, 22.8) (15, 3.3) (25, 10.4) (35, 7.7) (45, 6.5) (55, 4.2) (65, 7.1) (75, 5.6) (85, 5.9) (95, 26.4)
};
\end{axis}
    \end{tikzpicture}
    \caption{Original for Claude.}
\label{fig::meme-polarization_claude_orig}
\end{subfigure}%
\hfill%
\begin{subfigure}[b]{0.30\linewidth}
    \centering
    \begin{tikzpicture}
    \begin{axis}[
    scale only axis,
    width=\tonepanelw,
    height=\tonepanelh,
    ytick style={draw=none},
    xtick style={draw=none},
    ybar,
    bar width=3.5pt,
    xlabel={Tone Score},
    label style={font=\scriptsize},
    yticklabels=\empty,
    xmin=0, xmax=100,
    ymin=0, ymax=50,
    xtick={0,50,100},
    xticklabels={{\hspace{5pt}0},{50},{100\hspace{11pt}}},
    ytick={0,25,50},
    xticklabel style={
        font=\scriptsize
    },
    scaled y ticks=false,
    grid=major,
    legend style={at={(0.5,1.35)}, anchor=north, legend columns=-1, draw=none, nodes={inner sep=10pt}},
    legend image post style={
        xscale=0.5,
        yscale=1.5
    },
]
\addplot+[ybar,  fill=babyblue, draw=black, area legend, postaction={pattern=north west lines}] coordinates {
    (5, 7.0) (15, 3.6) (25, 12.3) (35, 11.1) (45, 15.9) (55, 12.5) (65, 11.3) (75, 9.7) (85, 7.0) (95, 9.5) 
};
\end{axis}
    \end{tikzpicture}
    \caption{\textit{Add} for Claude.}
\label{fig::meme-polarization_claude_add}
\end{subfigure}%
\hfill%
\begin{subfigure}[b]{0.30\linewidth}
    \centering
    \begin{tikzpicture}
    \begin{axis}[
    scale only axis,
    width=\tonepanelw,
    height=\tonepanelh,
    ytick style={draw=none},
    xtick style={draw=none},
    ybar,
    bar width=3.5pt,
    xlabel={Tone Score},
    label style={font=\scriptsize},
    yticklabels=\empty,
    xmin=0, xmax=100,
    ymin=0, ymax=50,
    xtick={0,50,100},
    xticklabels={{\hspace{5pt}0},{50},{100\hspace{11pt}}},
    ytick={0,25,50},
    xticklabel style={
        font=\scriptsize
    },
    scaled y ticks=false,
    grid=major,
    legend style={at={(0.5,1.35)}, anchor=north, legend columns=-1, draw=none, nodes={inner sep=10pt}},
    legend image post style={
        xscale=0.5,
        yscale=1.5
    },
]

\addplot+[ybar,  fill=babyred, draw=black, area legend, postaction={pattern=north east lines}] coordinates {
    (5, 38.9) (15, 0.0) (25, 13.2) (35, 0.0) (45, 0.0) (55, 0.1) (65, 1.1) (75, 6.8) (85, 4.7) (95, 35.3) 
};
\end{axis}
    \end{tikzpicture}
    \caption{\textit{Meme} for Claude.}
\label{fig::meme-polarization_claude_meme}
\end{subfigure}%
\hfill%
\begin{subfigure}[b]{0.37\linewidth}
    \centering
    \begin{tikzpicture}
    \begin{axis}[
    scale only axis,
    width=\tonepanelw,
    height=\tonepanelh,
    ytick style={draw=none},
    xtick style={draw=none},
    ybar,
    bar width=3.5pt,
    xlabel={Tone Score},
    label style={font=\scriptsize},
    ylabel={Share of Comments (\%)},
    xmin=0, xmax=100,
    ymin=0, ymax=50,
    xtick={0,50,100},
    xticklabels={{\hspace{5pt}0},{50},{100\hspace{11pt}}},
    ytick={0,25,50},
    xticklabel style={
        font=\scriptsize
    },
    yticklabel style={
        font=\scriptsize
    },
    scaled y ticks=false,
    grid=major,
    legend style={at={(0.5,1.35)}, anchor=north, legend columns=-1, draw=none, nodes={inner sep=10pt}},
    legend image post style={
        xscale=0.5,
        yscale=1.5
    },
]
\addplot+[ybar, bar shift=0pt, fill=gray!35, draw=black, area legend] coordinates {
    (5, 27.5) (15, 2.1) (25, 12.0) (35, 5.8) (45, 6.9) (55, 6.7) (65, 4.9) (75, 4.4) (85, 6.7) (95, 22.9) 
};
\end{axis}
    \end{tikzpicture}
    \caption{Original for DeepSeek.}
\label{fig::meme-polarization_ds_orig}
\end{subfigure}%
\hfill%
\begin{subfigure}[b]{0.30\linewidth}
    \centering
    \begin{tikzpicture}
    \begin{axis}[
    scale only axis,
    width=\tonepanelw,
    height=\tonepanelh,
    ytick style={draw=none},
    xtick style={draw=none},
    ybar,
    bar width=3.5pt,
    xlabel={Tone Score},
    label style={font=\scriptsize},
    yticklabels=\empty,
    xmin=0, xmax=100,
    ymin=0, ymax=50,
    xtick={0,50,100},
    xticklabels={{\hspace{5pt}0},{50},{100\hspace{11pt}}},
    ytick={0,25,50},
    xticklabel style={
        font=\scriptsize
    },
    scaled y ticks=false,
    grid=major,
    legend style={at={(0.5,1.35)}, anchor=north, legend columns=-1, draw=none, nodes={inner sep=10pt}},
    legend image post style={
        xscale=0.5,
        yscale=1.5
    },
]
\addplot+[ybar,  fill=babyblue, draw=black, area legend, postaction={pattern=north west lines}] coordinates {
    (5, 6.0) (15, 4.1) (25, 8.8) (35, 18.1) (45, 10.5) (55, 12.8) (65, 10.3) (75, 11.2) (85, 7.9) (95, 10.4) 
};
\end{axis}
    \end{tikzpicture}
    \caption{\textit{Add} for DeepSeek.}
\label{fig::meme-polarization_ds_add}
\end{subfigure}%
\hfill%
\begin{subfigure}[b]{0.30\linewidth}
    \centering
    \begin{tikzpicture}
    \begin{axis}[
    scale only axis,
    width=\tonepanelw,
    height=\tonepanelh,
    ytick style={draw=none},
    xtick style={draw=none},
    ybar,
    bar width=3.5pt,
    xlabel={Tone Score},
    label style={font=\scriptsize},
    yticklabels=\empty,
    xmin=0, xmax=100,
    ymin=0, ymax=50,
    xtick={0,50,100},
    xticklabels={{\hspace{5pt}0},{50},{100\hspace{11pt}}},
    ytick={0,25,50},
    xticklabel style={
        font=\scriptsize
    },
    scaled y ticks=false,
    grid=major,
    legend style={at={(0.5,1.35)}, anchor=north, legend columns=-1, draw=none, nodes={inner sep=10pt}},
    legend image post style={
        xscale=0.5,
        yscale=1.5
    },
]

\addplot+[ybar,  fill=babyred, draw=black, area legend, postaction={pattern=north east lines}] coordinates {
    (5, 46.0) (15, 0.0) (25, 12.4) (35, 0.0) (45, 0.0) (55, 0.1) (65, 1.2) (75, 4.5) (85, 4.7) (95, 31.1) 
};
\end{axis}
    \end{tikzpicture}
    \caption{\textit{Meme} for DeepSeek.}
\label{fig::meme-polarization_ds_meme}
\end{subfigure}%
\hfill%
\begin{subfigure}[b]{0.37\linewidth}
    \centering
    \begin{tikzpicture}
    \begin{axis}[
    scale only axis,
    width=\tonepanelw,
    height=\tonepanelh,
    ytick style={draw=none},
    xtick style={draw=none},
    ybar,
    bar width=3.5pt,
    xlabel={Tone Score},
    label style={font=\scriptsize},
    ylabel={Share of Comments (\%)},
    xmin=0, xmax=100,
    ymin=0, ymax=50,
    xtick={0,50,100},
    xticklabels={{\hspace{5pt}0},{50},{100\hspace{11pt}}},
    ytick={0,25,50},
    xticklabel style={
        font=\scriptsize
    },
    yticklabel style={
        font=\scriptsize
    },
    scaled y ticks=false,
    grid=major,
    legend style={at={(0.5,1.35)}, anchor=north, legend columns=-1, draw=none, nodes={inner sep=10pt}},
    legend image post style={
        xscale=0.5,
        yscale=1.5
    },
]
\addplot+[ybar, bar shift=0pt, fill=gray!35, draw=black, area legend] coordinates {
    (5, 28.5) (15, 1.3) (25, 11.5) (35, 3.2) (45, 4.5) (55, 5.4) (65, 3.2) (75, 4.5) (85, 6.1) (95, 31.7) 
};
\end{axis}
    \end{tikzpicture}
    \caption{Original for Llama.}
\label{fig::meme-polarization_llama_orig}
\end{subfigure}%
\hfill%
\begin{subfigure}[b]{0.30\linewidth}
    \centering
    \begin{tikzpicture}
    \begin{axis}[
    scale only axis,
    width=\tonepanelw,
    height=\tonepanelh,
    ytick style={draw=none},
    xtick style={draw=none},
    ybar,
    bar width=3.5pt,
    xlabel={Tone Score},
    label style={font=\scriptsize},
    yticklabels=\empty,
    xmin=0, xmax=100,
    ymin=0, ymax=50,
    xtick={0,50,100},
    xticklabels={{\hspace{5pt}0},{50},{100\hspace{11pt}}},
    ytick={0,25,50},
    xticklabel style={
        font=\scriptsize
    },
    scaled y ticks=false,
    grid=major,
    legend style={at={(0.5,1.35)}, anchor=north, legend columns=-1, draw=none, nodes={inner sep=10pt}},
    legend image post style={
        xscale=0.5,
        yscale=1.5
    },
]
\addplot+[ybar,  fill=babyblue, draw=black, area legend, postaction={pattern=north west lines}] coordinates {
    (5, 8.5) (15, 1.8) (25, 14.2) (35, 14.0) (45, 13.3) (55, 8.8) (65, 10.3) (75, 9.6) (85, 7.7) (95, 11.8) 
};
\end{axis}
    \end{tikzpicture}
    \caption{\textit{Add} for Llama.}
\label{fig::meme-polarization_llama_add}
\end{subfigure}%
\hfill%
\begin{subfigure}[b]{0.30\linewidth}
    \centering
    \begin{tikzpicture}
    \begin{axis}[
    scale only axis,
    width=\tonepanelw,
    height=\tonepanelh,
    ytick style={draw=none},
    xtick style={draw=none},
    ybar,
    bar width=3.5pt,
    xlabel={Tone Score},
    label style={font=\scriptsize},
    yticklabels=\empty,
    xmin=0, xmax=100,
    ymin=0, ymax=50,
    xtick={0,50,100},
    xticklabels={{\hspace{5pt}0},{50},{100\hspace{11pt}}},
    ytick={0,25,50},
    xticklabel style={
        font=\scriptsize
    },
    scaled y ticks=false,
    grid=major,
    legend style={at={(0.5,1.35)}, anchor=north, legend columns=-1, draw=none, nodes={inner sep=10pt}},
    legend image post style={
        xscale=0.5,
        yscale=1.5
    },
]

\addplot+[ybar,  fill=babyred, draw=black, area legend, postaction={pattern=north east lines}] coordinates {
    (5, 39.9) (15, 0.0) (25, 15.9) (35, 0.0) (45, 0.0) (55, 0.1) (65, 0.2) (75, 3.7) (85, 4.5) (95, 35.6) 
};
\end{axis}
    \end{tikzpicture}
    \caption{\textit{Meme} for Llama.}
\label{fig::meme-polarization_llama_meme}
\end{subfigure}
    \caption{Distribution of tone scores among original comments (gray), add-revised comments (blue), and meme-revised comments (red). Compared with the original comments, the \textit{add} strategy concentrates responses in the moderate tone range, while the \textit{meme} strategy shifts responses toward more extreme tone values. }
    \label{fig:meme-polarization_claude}
\end{figure}
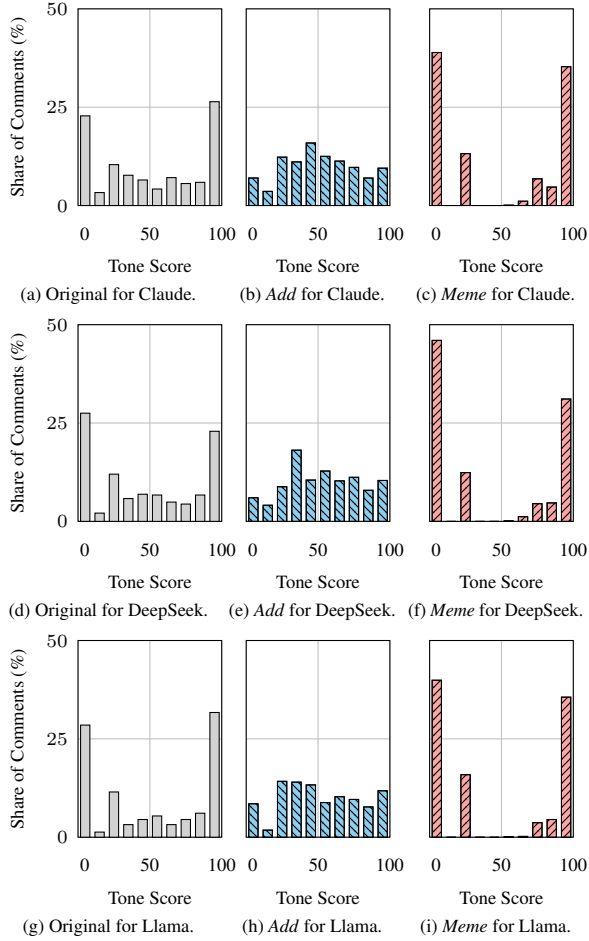

\subsection{Meme-Based Revision Introduces a Distinct Multimodal Effect}
\label{sec:results_meme}

To better understand why meme-based revisions are effective, we design four ablation variants that isolate the contribution of meme templates during the revision generation stage and the stance inference stage. Table~\ref{tab:meme_variant} summarizes the differences among these variants.

Specifically, \textit{r\_white\_meme} removes visual information during stance inference while keeping the generated meme text unchanged. The remaining variants replace the meme template during revision generation with progressively richer textual alternatives. \textit{r\_humor} instructs the revision model to generate humorous meme-style text without access to a meme template. \textit{r\_caption\_cut} replaces the meme image with a caption describing its visual content, while \textit{r\_caption} further augments the caption with external knowledge about the meme's common usage and cultural meaning. The results are shown in Table~\ref{tab:meme-ablation}.

\subsubsection{Are Meme Templates Useful During Stance Inference?}

To isolate the role of visual information during stance inference, we compare \textit{r\_meme} with \textit{r\_white\_meme}. Both variants use exactly the same generated meme text, which is produced from the same meme template. The only difference is in how the content is presented to the stance simulator:
\textit{r\_meme} retains the original meme template, whereas \textit{r\_white\_meme} replaces the meme image with a plain white background while preserving the generated text and its spatial layout.

We find that \textit{r\_meme} in general outperforms \textit{r\_white\_meme}. Since the textual content is identical between the two variants, this performance gap suggests that the visual information contained in the meme template provides additional cues during interpretation. The meme image therefore contributes more than a decorative presentation of the generated text. Instead, it supplies a complementary multimodal context that influences how the revised message is understood and ultimately affects the simulated user's stance.

Notably, this comparison isolates the effect of visual information during inference rather than throughout the entire pipeline. Because both variants use text generated from the same meme template, the observed difference indicates that meme images remain beneficial even after the revision text has already been generated.

\subsubsection{Are Meme Templates Useful During Revision Generation?}

We next investigate whether meme templates contribute useful information during the revision generation stage. To this end, we compare \textit{r\_meme} against \textit{r\_humor}, \textit{r\_caption\_cut}, and \textit{r\_caption}, all of which replace the meme template with textual alternatives when generating the revised content.

Across these variants, \textit{r\_meme} generally achieves the strongest stance-shifting effect. This finding suggests that the meme template provides information that is not fully captured by textual substitutes, even when those substitutes include explicit descriptions of the meme and external knowledge regarding its typical usage. While captions can describe visual content and usage instructions can summarize cultural meaning, they may fail to preserve subtle multimodal signals embedded in the original template. Such signals can include emotional expression, visual framing, implied speaker attitudes, relationships between image regions and text placement, and culturally grounded interpretations that emerge from the interaction between visual and textual elements.

Importantly, these results should not be interpreted as showing that images inherently contain more information than text. Rather, they indicate that the multimodal information contained in meme templates is more effectively utilized by the revision model than the textual approximations considered in our study. The superior performance of \textit{r\_meme} therefore suggests that directly conditioning on meme templates enables the revision model to generate more persuasive and contextually appropriate revisions than conditioning on humor instructions or caption-based descriptions alone.

Taken together, these findings reveal a dual role of meme templates within our framework. During revision generation, meme templates provide multimodal cues that cannot be fully replaced by textual approximations. During stance inference, the same templates offer complementary visual context that shapes how the revised message is interpreted. This dual contribution helps explain why meme-based revisions are consistently more effective than text-only alternatives in shifting simulated user stances.


\subsection{Revision Mechanisms Are Different Across Strategies}\label{sec:results_mechanism}
We further dive into comparing the differences between revised texts from the text-only and meme-based strategies. We use LIWC~\cite{boyd2022liwc22} to extract linguistic features from the texts generated by \textit{Add} and \textit{Meme} strategies and focus on the Tone feature, which indicates the overall emotional tone of a text, with higher scores indicating a more positive, upbeat, or optimistic style.

\begin{table}[t]
\centering
\resizebox{\columnwidth}{!}{
\begin{tabular}{@{}llccc@{}}
\toprule
\textbf{Model} & \textbf{\begin{tabular}[c]{@{}l@{}}Revision\\ Type\end{tabular}} & \textbf{\begin{tabular}[c]{@{}c@{}}Low-to-High\\ Shift (\%)\end{tabular}} & \textbf{\begin{tabular}[c]{@{}c@{}}High-to-Low\\ Shift (\%)\end{tabular}} & \textbf{\begin{tabular}[c]{@{}c@{}}Overall Depolarize\\ Rate (\%)\end{tabular}} \\ \midrule
DeepSeek & Add & 83.7 & 72.8 & 78.7 \\
DeepSeek & Meme & 47.8 & 68.8 & 56.6 \\ \midrule
Claude & Add & 81.0 & 82.3 & 81.7 \\
Claude & Meme & 52.0 & 66.4 & 58.6 \\ \midrule
Llama & Add & 81.2 & 77.7 & 79.4 \\
Llama & Meme & 47.2 & 62.8 & 54.5 \\ \bottomrule
\end{tabular}}
\caption{Tone-shift depolarization rates across revision types and models.}
\label{tab:meme-antipolar}
\end{table}

As shown in Figure~\ref{fig:meme-polarization_claude}, the Tone score distribution in original user comments exhibits a trend toward bipolar polarization. \textit{Add} significantly reduces this polarization, whereas \textit{meme} significantly intensifies it. Additionally, we take a directional depolarization statistical analysis. We refer to low-to-high Tone shift as cases where the original observed Tone score was no greater than 50 and increased after revision, whereas high-to-low shift as cases where the original Tone score was greater than 50 and decreased after revision. As shown in Table~\ref{tab:meme-antipolar}, \textit{add} shows a strong depolarizing trend in both directions, whereas \textit{meme} shifts the Tone of the text almost randomly. We also discover this trend in different subtopics, as shown in Appendix~\ref{app:topic}. This finding is consistent with persuasion theory~\citep{petty1986elaboration}: \textit{add} persuades by depolarizing the affective tone of the context, whereas \textit{meme} injects affective cues without a systematic tonal direction. We interpret this as a mechanism-level description of simulated behavior rather than a causal claim about persuasion in real users.

\subsection{Revision Effects Are Robust Across Simulators and Domains}
\label{sec:results_robustness}

Finally, we evaluate whether revision effects are robust across revision models and stance models (simulators). Appendix~\ref{app:robust} reports the detailed average directional stance shift and stance transition rates under different settings. Additionally, we conduct sensitive analysis over different temperatures, meme templates, and prompt variants in Appendix~\ref{app:sensitive} to evaluate the internal robustness of LLMs. The results show that the main ranking of revision strategies is largely consistent across stance models. High consistency is observed across prompts and temperatures as well.
Beyond simulator robustness, we further test whether the revision effects generalize beyond AI-related discussions by replicating the full pipeline on four new domains with four new stance targets: nuclear energy, the keto diet, universal basic income, and the SpaceX IPO, using 1,200 additional Reddit conversations. The cross-domain averages closely align with Table~\ref{tab:main-delta}, preserving the same strategy ordering, with the average differences  significant at $p<0.001$.
The tone-based mechanism signatures identified in Section~\ref{sec:results_mechanism} also replicate across domains. Full results are reported in Appendix~\ref{app:domains}.

\section{Related Work}
LLMs are increasingly used as proxies for human respondents and online users from ``silicon samples''~\citep{argyle2023out} and generative agents~\citep{park2023generative} to opinion-dynamics simulation~\citep{chuang2024simulating,zhang2024electionsim,zhang2025socioverse}, and more broadly as zero-shot tools across the CSS pipeline~\citep{ziems2024can, gilardi2023chatgpt}. A related thread is stance detection~\citep{kucuk2020stance}, now often tackled by prompting or adapting LLMs~\citep{zhang2024llm, zhao2024ezstance}, with memes studied as a multimodal stance channel~\citep{kiela2020hateful, 10.1145/3709005}. However, LLM opinion outputs are known to be unstable under superficial prompt-format~\citep{sclar2024quantifying} and questionnaire-design~\citep{rottger2024political} changes. Unlike this prior auditing of prompt surface form, we audit sensitivity to controlled, plausibility-constrained revisions of the conversational context, and extend it to the multimodal setting, characterizing whether simulated stance shifts and why.

\section{Conclusion}
This study audits LLM-based stance simulation in online discussions through counterfactual revision strategies and extensive analysis. 
Our findings reveal robust and effective stance transition improvements through several strategies, along with distinct revision mechanisms.

\section*{Limitations}
This study has several limitations.
First, our main analysis focuses on online discussions about different LLMs (Claude, Deepseek, and Llama), with all conversations collected from Reddit. Although the extension in Appendix~\ref{app:domains} replicates the main findings across four additional domains, the data remain limited to a single platform. Extending our framework to other platforms would require comparably structured conversations data and represents an important direction for future work. Second, our audit is designed to evaluate the context sensitivity of LLM-simulated stances and does not establish whether these simulated shifts correspond to real human opinion change. Validating such correspondence would require comparing simulated stance shifts with observed human responses, which we leave for future work.

\section*{Ethical Considerations}

This study evaluates how LLM-based stance simulations respond to revised social contexts. It does not aim to develop methods for manipulating real users or changing real public opinion. All revision strategies are used as controlled interventions for auditing model behavior. We therefore interpret stance shifts as changes in LLM simulated stance, not as evidence of actual human opinion change.

Because Reddit data may contain personal or sensitive information, we remove or anonymize user identifiers and avoid reporting examples that could reveal user identity. We also avoid generating revised contexts that contain harassment, private information, or harmful misinformation. When presenting qualitative examples, we paraphrase or mask identifying details where necessary. These precautions are important because the goal of the study is to understand the reliability and risks of LLM-based social simulation, rather than to reproduce or amplify harmful online content.

We also note the dual-use nature of this framework. The revision strategies that we employ as controlled probes could, in principle, be repurposed to craft persuasive interventions or fabricate synthetic public opinion at scale. To mitigate this, we do not release any component optimized for persuasion beyond the auditing pipeline itself, and we believe that systematically characterizing the context sensitivity of LLM-based user simulations is a prerequisite for mitigating such misuse, as it reveals when and how simulated opinions are susceptible to contextual manipulation.

\section*{Acknowledgments}
We use an AI assistant for language editing and for generating certain visual elements during manuscript preparation. All scientific figures and plots are created by the authors.

\bibliography{custom}

\appendix

\section{Additional Details of Data Collection and Preprocessing}\label{appendix_sec:data}

Understanding how users respond to evolving technological systems is important because online discussions increasingly shape public perception, technology adoption, trust, and acceptance. Discussions surrounding new AI systems are particularly suitable for this purpose, as they often involve rapidly changing information, strong community engagement, competing narratives, and diverse forms of evidence ranging from technical evaluations to memes and social commentary. These characteristics make them a useful case study for examining how conversational context influences simulated user stance.

For each model family, we collect conversations that mention the model itself, its developer or company, and closely related topics, including model capability, release events, open source availability, safety concerns, pricing, accessibility, corporate reputation, and comparisons with other LLMs. The initial search ranges across 443 unique tech-related subreddits, where each subreddit represents a distinct Reddit community with its own discussion norms and user interests. The broad search scope allows our dataset to capture diverse technical topics across the current LLM ecosystem. This collection strategy allows us to study stance simulation across multiple targets that differ in public perception, technical framing, and community discussion patterns.

Each data instance is constructed from a Reddit conversation thread. We retain the post title, post body, comment structure, comment text, timestamps, subreddit information, and the conversational path leading to a target user's comment. We first remove conversations containing deleted comments that appear in the middle because these interruptions can break the conversational flow and make stance interpretation unreliable. Keeping only conversations in which the target user speaks at least twice ensures there is sufficient prior context for reliable stance inference. In the whole conversation context, the target user's final comment is used to validate the observed stance, while the preceding conversational context is used as the input for stance simulation. 

We define the stance target at the level of an LLM family or its associated organization. For example, a discussion may concern DeepSeek as a model, DeepSeek as a company, or broader issues associated with DeepSeek, such as open source competitiveness or geopolitical concerns. Similarly, Claude-related discussions may involve the model, Anthropic, safety positioning, pricing, or user experience; Llama-related discussions may involve the model family, Meta, open source release strategies, or comparisons with proprietary models. To ensure the conversation is about the LLM family, we retain only instances in which the target user explicitly mentions the LLM name in their own comments. We further validate this filtering strategy through manual annotation by labeling whether conversations are substantively discussing the target model or its related context. Among 50 sampled conversations containing the selected keywords, 45 were judged to be relevant to the target model, compared to only 6 out of 50 conversations without the keywords. Based on this validation, we restrict our analysis to conversations where the target user’s text contains the selected model-related keywords. 

During preprocessing, we identify the primary stance target for each conversation and remove cases where the target is ambiguous or where the conversation does not contain sufficient contextual information for stance inference.

\section{Implementation Details}\label{appendix_sec:implementation}
During the stance simulation, we employ \texttt{gpt-5.2-2025-12-11}, \texttt{qwen3.5-plus-2026-02-15}, and \texttt{claude-sonnet-4-6} as the stance models to infer the target user's stance and verify the robustness of our study setup~\cite{openai2025gpt52system,qwen2026qwen35modelcard,anthropic2026sonnet46}. We report results mainly based on GPT-5.2 due to space restrictions, with the full scale results in Appendix~\ref{subapp:main-full}.

For text revision, we use both \texttt{gemini-3-flash-preview} and \texttt{claude-haiku-4-5-20251001} as revision models. To avoid potential self-preference or model-specific bias, \texttt{claude-haiku-4-5-20251001} is not used to revise conversations related to Claude~\cite{google2025gemini3models,anthropic2025haiku45}. Due to space limitations, the main text reports results based on Gemini revisions only, while the Claude revision results are provided in Appendix~\ref{appendix:claude_result}

During the multimodal revision, we employ only \texttt{gemini-3-flash-preview} as the revision model to generate the meme text and \texttt{gpt-image-2} to generate the final meme~\cite{openai2026gptimage2}.

\section{Human Validation of Revised Contexts}\label{app:human-validation}

To verify that the counterfactual revisions are valid, semantically controlled interventions, we conduct a human judgment study on the revised contexts.

\paragraph{Setup.} Three annotators independently evaluate 60 revised contexts each (15 conversations covering all three stance targets $\times$ 4 revision strategies), yielding 540 binary judgments in total. For each instance, annotators see the full original conversation thread, the original last other-user message, and the revision produced by each strategy, with memes shown as full images. Each revision is judged on three binary criteria: (a) \textbf{strategy adherence}: the revision faithfully implements its intended strategy; (b) \textbf{coherence}: the revision is coherent with the conversational context; and (c) \textbf{target preservation}: the revision preserves the original stance target and core argument without concept switching.

\begin{table}[t]
\centering
\resizebox{0.9\linewidth}{!}{
\begin{tabular}{@{}lcc@{}}
\toprule
\textbf{Criterion} & \textbf{Pass Rate} & \textbf{IAA} \\ \midrule
Strategy adherence & 92.8\% & 86.7\% \\
Coherence & 95.6\% & 93.3\% \\
Target preservation & 100.0\% & 100.0\% \\ \bottomrule
\end{tabular}
}
\caption{Human validation of the revised contexts: pass rate and average pairwise inter-annotator agreement (IAA) on each criterion.}
\label{tab:human-validation}
\end{table}

\paragraph{Results.} Table~\ref{tab:human-validation} reports the pass rate and the inter-annotator agreement for each criterion. Per strategy, the all-criteria pass rate is 100\% for \textit{paraphrase}, 93.3\% for \textit{add}, 86.7\% for \textit{meme}, and 77.8\% for \textit{explain}. These results indicate that, by human judgment, the revisions implement the intended contextual change while keeping the stance target fixed. The stance transitions we report can therefore be interpreted as the simulator's response to human-verified, meaning-preserving context interventions, rather than to uncontrolled LLM generation noise or a self-referential loop between the revision model and the stance simulator.

\section{Topic-Level Consistency of Polarization Trends}\label{app:topic}

To further examine whether these polarization patterns are driven solely by differences in discussion content, following \citet{QI2024102158}, we conduct a topic-level analysis of the revised conversations. Topic modeling is performed using BERTopic~\cite{grootendorst2022bertopic} on the original conversations to identify semantically similar discussion themes. Conversation embeddings are generated using the \texttt{all-MiniLM-L6-v2} Sentence-Transformers model, followed by UMAP dimensionality reduction and HDBSCAN clustering to identify semantic topics~\cite{mcinnes2018umap, wang2020minilm, mcinnes2017hdbscan}.

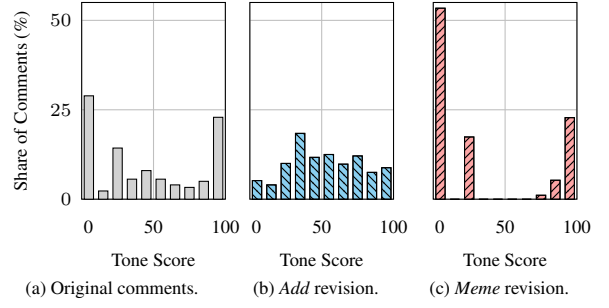
\begin{figure}[t]
    \centering
    \captionsetup[subfigure]{font=scriptsize,skip=1pt}
    \begin{subfigure}[b]{0.37\linewidth}
    \centering
    \begin{tikzpicture}
    \begin{axis}[
    scale only axis,
    width=\tonepanelw,
    height=\tonepanelh,
    ytick style={draw=none},
    xtick style={draw=none},
    ybar,
    bar width=3.5pt,
    xlabel={Tone Score},
    label style={font=\scriptsize},
    ylabel={Share of Comments (\%)},
    xmin=0, xmax=100,
    ymin=0, ymax=55,
    xtick={0,50,100},
    xticklabels={{\hspace{5pt}0},{50},{100\hspace{11pt}}},
    ytick={0,25,50},
    xticklabel style={
        font=\scriptsize
    },
    yticklabel style={
        font=\scriptsize
    },
    scaled y ticks=false,
    grid=major,
    legend style={at={(0.5,1.35)}, anchor=north, legend columns=-1, draw=none, nodes={inner sep=10pt}},
    legend image post style={
        xscale=0.5,
        yscale=1.5
    },
]
\addplot+[ybar, bar shift=0pt, fill=gray!35, draw=black, area legend] coordinates {
    (5, 28.9) (15, 2.3) (25, 14.3) (35, 5.6) (45, 8.0) (55, 5.6) (65, 4.0) (75, 3.3) (85, 5.0) (95, 22.9) 
};
\end{axis}
    \end{tikzpicture}
    \caption{Original comments.}
\label{fig::meme-polarization_ai_model_evaluation_orig}
\end{subfigure}%
\hfill%
\begin{subfigure}[b]{0.30\linewidth}
    \centering
    \begin{tikzpicture}
    \begin{axis}[
    scale only axis,
    width=\tonepanelw,
    height=\tonepanelh,
    ytick style={draw=none},
    xtick style={draw=none},
    ybar,
    bar width=3.5pt,
    xlabel={Tone Score},
    label style={font=\scriptsize},
    yticklabels=\empty,
    xmin=0, xmax=100,
    ymin=0, ymax=55,
    xtick={0,50,100},
    xticklabels={{\hspace{5pt}0},{50},{100\hspace{11pt}}},
    ytick={0,25,50},
    xticklabel style={
        font=\scriptsize
    },
    scaled y ticks=false,
    grid=major,
    legend style={at={(0.5,1.35)}, anchor=north, legend columns=-1, draw=none, nodes={inner sep=10pt}},
    legend image post style={
        xscale=0.5,
        yscale=1.5
    },
]
\addplot+[ybar,  fill=babyblue, draw=black, area legend, postaction={pattern=north west lines}] coordinates {
    (5, 5.2) (15, 4.0) (25, 10.0) (35, 18.4) (45, 11.7) (55, 12.5) (65, 9.8) (75, 12.1) (85, 7.5) (95, 8.8)
};
\end{axis}
    \end{tikzpicture}
    \caption{\textit{Add} revision.}
\label{fig::meme-polarization_ai_model_evaluation_add}
\end{subfigure}%
\hfill%
\begin{subfigure}[b]{0.30\linewidth}
    \centering
    \begin{tikzpicture}
    \begin{axis}[
    scale only axis,
    width=\tonepanelw,
    height=\tonepanelh,
    ytick style={draw=none},
    xtick style={draw=none},
    ybar,
    bar width=3.5pt,
    xlabel={Tone Score},
    label style={font=\scriptsize},
    yticklabels=\empty,
    xmin=0, xmax=100,
    ymin=0, ymax=55,
    xtick={0,50,100},
    xticklabels={{\hspace{5pt}0},{50},{100\hspace{11pt}}},
    ytick={0,25,50},
    xticklabel style={
        font=\scriptsize
    },
    scaled y ticks=false,
    grid=major,
    legend style={at={(0.5,1.35)}, anchor=north, legend columns=-1, draw=none, nodes={inner sep=10pt}},
    legend image post style={
        xscale=0.5,
        yscale=1.5
    },
]

\addplot+[ybar,  fill=babyred, draw=black, area legend, postaction={pattern=north east lines}] coordinates {
    (5, 53.4) (15, 0.0) (25, 17.4) (35, 0.0) (45, 0.0) (55, 0.0) (65, 0.0) (75, 1.1) (85, 5.3) (95, 22.8)
};
\end{axis}
    \end{tikzpicture}
    \caption{\textit{Meme} revision.}
\label{fig::meme-polarization_ai_model_evaluation_meme}
\end{subfigure}
    \caption{Distribution of tone scores among the original comments (gray), add-revised comments (blue), and meme-revised comments (red) of the \textit{AI Model Evaluation} topic.}
    \label{fig:meme-polarization_ai_model_evaluation}
\end{figure}

\begin{figure}[t]
    \centering
    \captionsetup[subfigure]{font=scriptsize,skip=1pt}
    \begin{subfigure}[b]{0.37\linewidth}
    \centering
    \begin{tikzpicture}
    \begin{axis}[
    scale only axis,
    width=\tonepanelw,
    height=\tonepanelh,
    ytick style={draw=none},
    xtick style={draw=none},
    ybar,
    bar width=3.5pt,
    xlabel={Tone Score},
    label style={font=\scriptsize},
    ylabel={Share of Comments (\%)},
    xmin=0, xmax=100,
    ymin=0, ymax=50,
    xtick={0,50,100},
    xticklabels={{\hspace{5pt}0},{50},{100\hspace{11pt}}},
    ytick={0,25,50},
    xticklabel style={
        font=\scriptsize
    },
    yticklabel style={
        font=\scriptsize
    },
    scaled y ticks=false,
    grid=major,
    legend style={at={(0.5,1.35)}, anchor=north, legend columns=-1, draw=none, nodes={inner sep=10pt}},
    legend image post style={
        xscale=0.5,
        yscale=1.5
    },
]
\addplot+[ybar, bar shift=0pt, fill=gray!35, draw=black, area legend] coordinates {
    (5, 17.6) (15, 2.1) (25, 4.2) (35, 4.9) (45, 6.3) (55, 4.9) (65, 8.5) (75, 8.5) (85, 9.9) (95, 33.1) 
};
\end{axis}
    \end{tikzpicture}
    \caption{Original comments.}
\label{fig::meme-polarization_software_orig}
\end{subfigure}%
\hfill%
\begin{subfigure}[b]{0.30\linewidth}
    \centering
    \begin{tikzpicture}
    \begin{axis}[
    scale only axis,
    width=\tonepanelw,
    height=\tonepanelh,
    ytick style={draw=none},
    xtick style={draw=none},
    ybar,
    bar width=3.5pt,
    xlabel={Tone Score},
    label style={font=\scriptsize},
    yticklabels=\empty,
    xmin=0, xmax=100,
    ymin=0, ymax=50,
    xtick={0,50,100},
    xticklabels={{\hspace{5pt}0},{50},{100\hspace{11pt}}},
    ytick={0,25,50},
    xticklabel style={
        font=\scriptsize
    },
    scaled y ticks=false,
    grid=major,
    legend style={at={(0.5,1.35)}, anchor=north, legend columns=-1, draw=none, nodes={inner sep=10pt}},
    legend image post style={
        xscale=0.5,
        yscale=1.5
    },
]
\addplot+[ybar,  fill=babyblue, draw=black, area legend, postaction={pattern=north west lines}] coordinates {
    (5, 5.2) (15, 4.7) (25, 11.8) (35, 11.8) (45, 14.7) (55, 12.8) (65, 10.0) (75, 9.0) (85, 9.5) (95, 10.4) 
};
\end{axis}
    \end{tikzpicture}
    \caption{\textit{Add} revision.}
\label{fig::meme-polarization_software_add}
\end{subfigure}%
\hfill%
\begin{subfigure}[b]{0.30\linewidth}
    \centering
    \begin{tikzpicture}
    \begin{axis}[
    scale only axis,
    width=\tonepanelw,
    height=\tonepanelh,
    ytick style={draw=none},
    xtick style={draw=none},
    ybar,
    bar width=3.5pt,
    xlabel={Tone Score},
    label style={font=\scriptsize},
    yticklabels=\empty,
    xmin=0, xmax=100,
    ymin=0, ymax=50,
    xtick={0,50,100},
    xticklabels={{\hspace{5pt}0},{50},{100\hspace{11pt}}},
    ytick={0,25,50},
    xticklabel style={
        font=\scriptsize
    },
    scaled y ticks=false,
    grid=major,
    legend style={at={(0.5,1.35)}, anchor=north, legend columns=-1, draw=none, nodes={inner sep=10pt}},
    legend image post style={
        xscale=0.5,
        yscale=1.5
    },
]

\addplot+[ybar,  fill=babyred, draw=black, area legend, postaction={pattern=north east lines}] coordinates {
    (5, 43.7) (15, 0.0) (25, 12.6) (35, 0.0) (45, 0.0) (55, 0.0) (65, 0.0) (75, 0.8) (85, 7.6) (95, 35.3) 
};
\end{axis}
    \end{tikzpicture}
    \caption{\textit{Meme} revision.}
\label{fig::meme-polarization_software_meme}
\end{subfigure}
    \caption{Distribution of tone scores among the original comments (gray), add-revised comments (blue), and meme-revised comments (red) of the \textit{AI-Assisted Software Development Practices} topic.}
    \label{fig:meme-polarization_software}
\end{figure}
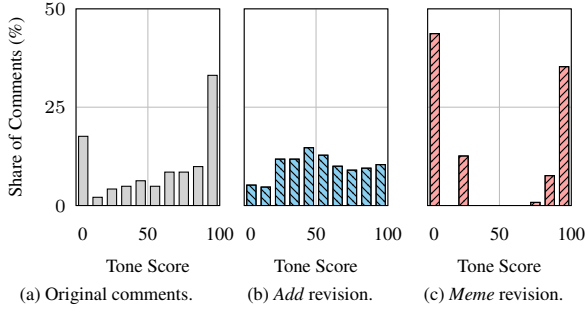

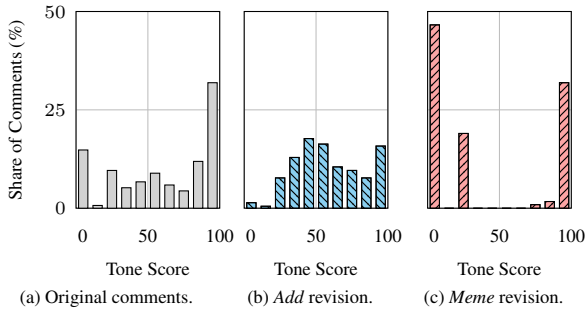
\begin{figure}[t]
    \centering
    \captionsetup[subfigure]{font=scriptsize,skip=1pt}
    \begin{subfigure}[b]{0.37\linewidth}
    \centering
    \begin{tikzpicture}
    \begin{axis}[
    scale only axis,
    width=\tonepanelw,
    height=\tonepanelh,
    ytick style={draw=none},
    xtick style={draw=none},
    ybar,
    bar width=3.5pt,
    xlabel={Tone Score},
    label style={font=\scriptsize},
    ylabel={Share of Comments (\%)},
    xmin=0, xmax=100,
    ymin=0, ymax=50,
    xtick={0,50,100},
    xticklabels={{\hspace{5pt}0},{50},{100\hspace{11pt}}},
    ytick={0,25,50},
    xticklabel style={
        font=\scriptsize
    },
    yticklabel style={
        font=\scriptsize
    },
    scaled y ticks=false,
    grid=major,
    legend style={at={(0.5,1.35)}, anchor=north, legend columns=-1, draw=none, nodes={inner sep=10pt}},
    legend image post style={
        xscale=0.5,
        yscale=1.5
    },
]
\addplot+[ybar, bar shift=0pt, fill=gray!35, draw=black, area legend] coordinates {
    (5, 14.8) (15, 0.7) (25, 9.6) (35, 5.2) (45, 6.7) (55, 8.9) (65, 5.9) (75, 4.4) (85, 11.9) (95, 31.9) 
};
\end{axis}
    \end{tikzpicture}
    \caption{Original comments.}
\label{fig::meme-polarization_cost_orig}
\end{subfigure}%
\hfill%
\begin{subfigure}[b]{0.30\linewidth}
    \centering
    \begin{tikzpicture}
    \begin{axis}[
    scale only axis,
    width=\tonepanelw,
    height=\tonepanelh,
    ytick style={draw=none},
    xtick style={draw=none},
    ybar,
    bar width=3.5pt,
    xlabel={Tone Score},
    label style={font=\scriptsize},
    yticklabels=\empty,
    xmin=0, xmax=100,
    ymin=0, ymax=50,
    xtick={0,50,100},
    xticklabels={{\hspace{5pt}0},{50},{100\hspace{11pt}}},
    ytick={0,25,50},
    xticklabel style={
        font=\scriptsize
    },
    scaled y ticks=false,
    grid=major,
    legend style={at={(0.5,1.35)}, anchor=north, legend columns=-1, draw=none, nodes={inner sep=10pt}},
    legend image post style={
        xscale=0.5,
        yscale=1.5
    },
]
\addplot+[ybar,  fill=babyblue, draw=black, area legend, postaction={pattern=north west lines}] coordinates {
   (5, 1.4) (15, 0.5) (25, 7.7) (35, 12.9) (45, 17.7) (55, 16.3) (65, 10.5) (75, 9.6) (85, 7.7) (95, 15.8) 
};
\end{axis}
    \end{tikzpicture}
    \caption{\textit{Add} revision.}
\label{fig::meme-polarization_cost_add}
\end{subfigure}%
\hfill%
\begin{subfigure}[b]{0.30\linewidth}
    \centering
    \begin{tikzpicture}
    \begin{axis}[
    scale only axis,
    width=\tonepanelw,
    height=\tonepanelh,
    ytick style={draw=none},
    xtick style={draw=none},
    ybar,
    bar width=3.5pt,
    xlabel={Tone Score},
    label style={font=\scriptsize},
    yticklabels=\empty,
    xmin=0, xmax=100,
    ymin=0, ymax=50,
    xtick={0,50,100},
    xticklabels={{\hspace{5pt}0},{50},{100\hspace{11pt}}},
    ytick={0,25,50},
    xticklabel style={
        font=\scriptsize
    },
    scaled y ticks=false,
    grid=major,
    legend style={at={(0.5,1.35)}, anchor=north, legend columns=-1, draw=none, nodes={inner sep=10pt}},
    legend image post style={
        xscale=0.5,
        yscale=1.5
    },
]

\addplot+[ybar,  fill=babyred, draw=black, area legend, postaction={pattern=north east lines}] coordinates {
    (5, 46.6) (15, 0.0) (25, 19.0) (35, 0.0) (45, 0.0) (55, 0.0) (65, 0.0) (75, 0.9) (85, 1.7) (95, 31.9) 
};
\end{axis}
    \end{tikzpicture}
    \caption{\textit{Meme} revision.}
\label{fig::meme-polarization_cost_meme}
\end{subfigure}
    \caption{Distribution of tone scores among the original comments (gray), add-revised comments (blue), and meme-revised comments (red) of the \textit{AI Model Training Costs and Efficiency} topic.}
    \label{fig:meme-polarization_cost}
\end{figure}

\begin{figure}[t]
    \centering
    \captionsetup[subfigure]{font=scriptsize,skip=1pt}
    \begin{subfigure}[b]{0.37\linewidth}
    \centering
    \begin{tikzpicture}
    \begin{axis}[
    scale only axis,
    width=\tonepanelw,
    height=\tonepanelh,
    ytick style={draw=none},
    xtick style={draw=none},
    ybar,
    bar width=3.5pt,
    xlabel={Tone Score},
    label style={font=\scriptsize},
    ylabel={Share of Comments (\%)},
    xmin=0, xmax=100,
    ymin=0, ymax=55,
    xtick={0,50,100},
    xticklabels={{\hspace{5pt}0},{50},{100\hspace{11pt}}},
    ytick={0,25,50},
    xticklabel style={
        font=\scriptsize
    },
    yticklabel style={
        font=\scriptsize
    },
    scaled y ticks=false,
    grid=major,
    legend style={at={(0.5,1.35)}, anchor=north, legend columns=-1, draw=none, nodes={inner sep=10pt}},
    legend image post style={
        xscale=0.5,
        yscale=1.5
    },
]
\addplot+[ybar, bar shift=0pt, fill=gray!35, draw=black, area legend] coordinates {
    (5, 34.5) (15, 3.6) (25, 9.1) (35, 5.5) (45, 7.3) (55, 5.5) (65, 10.9) (75, 1.8) (85, 3.6) (95, 18.2) 
};
\end{axis}
    \end{tikzpicture}
    \caption{Original comments.}
\label{fig::meme-polarization_consciousness_orig}
\end{subfigure}%
\hfill%
\begin{subfigure}[b]{0.30\linewidth}
    \centering
    \begin{tikzpicture}
    \begin{axis}[
    scale only axis,
    width=\tonepanelw,
    height=\tonepanelh,
    ytick style={draw=none},
    xtick style={draw=none},
    ybar,
    bar width=3.5pt,
    xlabel={Tone Score},
    label style={font=\scriptsize},
    yticklabels=\empty,
    xmin=0, xmax=100,
    ymin=0, ymax=55,
    xtick={0,50,100},
    xticklabels={{\hspace{5pt}0},{50},{100\hspace{11pt}}},
    ytick={0,25,50},
    xticklabel style={
        font=\scriptsize
    },
    scaled y ticks=false,
    grid=major,
    legend style={at={(0.5,1.35)}, anchor=north, legend columns=-1, draw=none, nodes={inner sep=10pt}},
    legend image post style={
        xscale=0.5,
        yscale=1.5
    },
]
\addplot+[ybar,  fill=babyblue, draw=black, area legend, postaction={pattern=north west lines}] coordinates {
    (5, 9.8) (15, 4.3) (25, 17.4) (35, 12.0) (45, 16.3) (55, 8.7) (65, 12.0) (75, 6.5) (85, 3.3) (95, 9.8) 
};
\end{axis}
    \end{tikzpicture}
    \caption{\textit{Add} revision.}
\label{fig::meme-polarization_consciousness_add}
\end{subfigure}%
\hfill%
\begin{subfigure}[b]{0.30\linewidth}
    \centering
    \begin{tikzpicture}
    \begin{axis}[
    scale only axis,
    width=\tonepanelw,
    height=\tonepanelh,
    ytick style={draw=none},
    xtick style={draw=none},
    ybar,
    bar width=3.5pt,
    xlabel={Tone Score},
    label style={font=\scriptsize},
    yticklabels=\empty,
    xmin=0, xmax=100,
    ymin=0, ymax=55,
    xtick={0,50,100},
    xticklabels={{\hspace{5pt}0},{50},{100\hspace{11pt}}},
    ytick={0,25,50},
    xticklabel style={
        font=\scriptsize
    },
    scaled y ticks=false,
    grid=major,
    legend style={at={(0.5,1.35)}, anchor=north, legend columns=-1, draw=none, nodes={inner sep=10pt}},
    legend image post style={
        xscale=0.5,
        yscale=1.5
    },
]

\addplot+[ybar,  fill=babyred, draw=black, area legend, postaction={pattern=north east lines}] coordinates {
   (5, 50.0) (15, 0.0) (25, 17.6) (35, 0.0) (45, 0.0) (55, 0.0) (65, 0.0) (75, 0.0) (85, 2.7) (95, 29.7) 
};
\end{axis}
    \end{tikzpicture}
    \caption{\textit{Meme} revision.}
\label{fig::meme-polarization_consciousness_meme}
\end{subfigure}
    \caption{Distribution of tone scores among the original comments (gray), add-revised comments (blue), and meme-revised comments (red) of the \textit{AI Language Models and Consciousness Debate} topic.}
    \label{fig:meme-polarization_consciousness}
\end{figure}
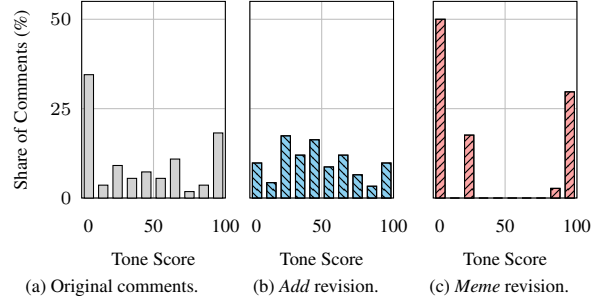

As shown in Figures~\ref{fig:meme-polarization_ai_model_evaluation}, \ref{fig:meme-polarization_software}, \ref{fig:meme-polarization_cost}, and \ref{fig:meme-polarization_consciousness}, the overall trend remains highly consistent across topics. This further supports the claim that \textit{add} demonstrates a depolarizing tendency and \textit{meme} amplifies polarization across discussion contexts.

\section{Robustness Verification of LLM-based Stance Simulation}\label{app:robust}
\subsection{Results on Three Stance Models}\label{subapp:main-full}
\begin{figure*}[t]
    \centering
    \includegraphics[width=0.8\linewidth]{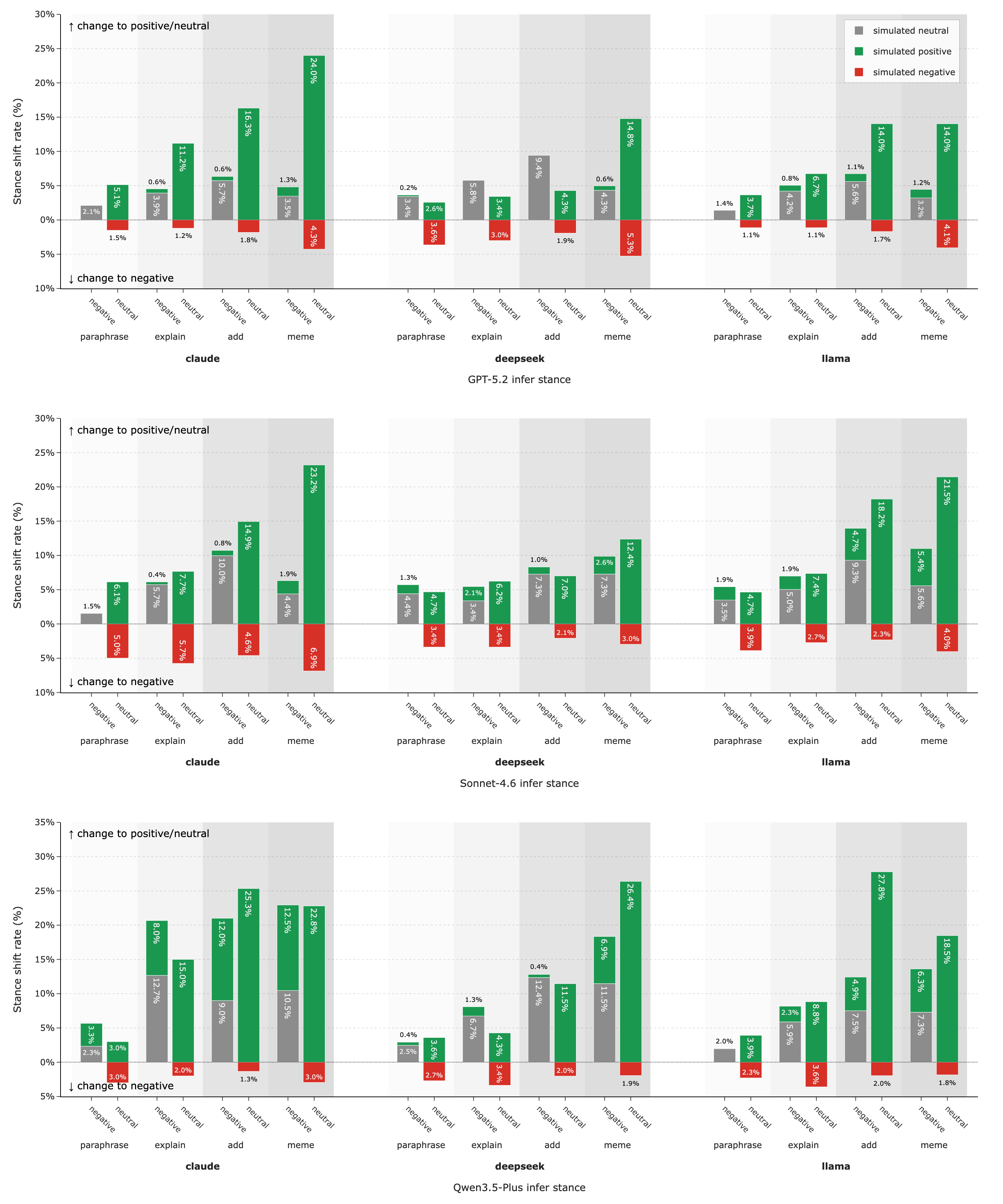}
    \caption{An illustration of the stance transition rates of different strategies on all target topics across three different stance models.}
    \label{fig:app-3-models}
\end{figure*}

Table~\ref{tab:app-main-delta} and Figure~\ref{fig:app-3-models} show the full-scale experiment results in both metrics across different stance models (GPT-5.2, Sonnet-4.6, and Qwen3.5), with Gemini as the revision model. \textit{meme} remains among the most effective strategies across models, suggesting that its effect is not limited to a single LLM. In contrast, \textit{add} shows more model-specific variation, indicating that some revision effects may depend on the simulator's own priors, instruction following behavior, or sensitivity to conversational framing.

\begin{table*}[t]
\centering
\small
\setlength{\tabcolsep}{4pt}
\begin{tabular}{@{}lccccc@{}}
\toprule
{\textbf{Target Topic}} & \textbf{Inferred} & \textbf{paraphrase \(\Delta\)} & \textbf{explain \(\Delta\)} & \textbf{add \(\Delta\)} & \textbf{meme \(\Delta\)} \\ \midrule
\multicolumn{6}{c}{GPT-5.2-Inferred Stance} \\ \midrule
Claude & 0.234 & +0.002 (+0.8\%) & +0.056 (+23.8\%) & +0.113 (+48.4\%) & +0.141 (+60.3\%) \\
Deepseek & 0.170 & -0.011 (-6.7\%) & -0.008 (-4.5\%) & +0.029 (+17.2\%) & +0.076 (+44.4\%) \\
Llama & 0.133 & -0.010 (-7.6\%) & +0.067 (+50.0\%) & +0.125 (+93.9\%) & +0.051 (+38.6\%) \\ \midrule
\multicolumn{6}{c}{Sonnet-4.6-Inferred Stance} \\ \midrule
Claude & 0.339 & -0.024 (-7.2\%) & -0.004 (-1.1\%) & +0.079 (+23.2\%) & +0.082 (+24.1\%) \\
Deepseek & 0.249 & +0.011 (+4.6\%) & +0.013 (+5.1\%) & +0.037 (+14.8\%) & +0.066 (+26.7\%) \\
Llama & 0.269 & 0.000 (+0.0\%) & +0.012 (+4.6\%) & +0.167 (+62.3\%) & +0.129 (+48.1\%) \\ \midrule
\multicolumn{6}{c}{Qwen3.5-Plus-Inferred Stance} \\ \midrule
Claude & 0.186 & +0.015 (+8.0\%) & +0.205 (+110.0\%) & +0.302 (+162.0\%) & +0.288 (+154.8\%) \\
Deepseek & 0.154 & -0.012 (-7.6\%) & +0.010 (+6.7\%) & +0.092 (+59.7\%) & +0.266 (+173.1\%) \\
Llama & 0.163 & -0.010 (-6.2\%) & +0.050 (+30.9\%) & +0.252 (+154.3\%) & +0.134 (+82.0\%) \\ \bottomrule
\end{tabular}
\caption{The average directional stance shift of different strategies on all target topics and stance models.}
\label{tab:app-main-delta}
\end{table*}

\subsection{Results on Claude-Based Revision Strategy}
\label{appendix:claude_result}

\begin{figure*}[t]
    \centering
    \includegraphics[width=0.9\linewidth]{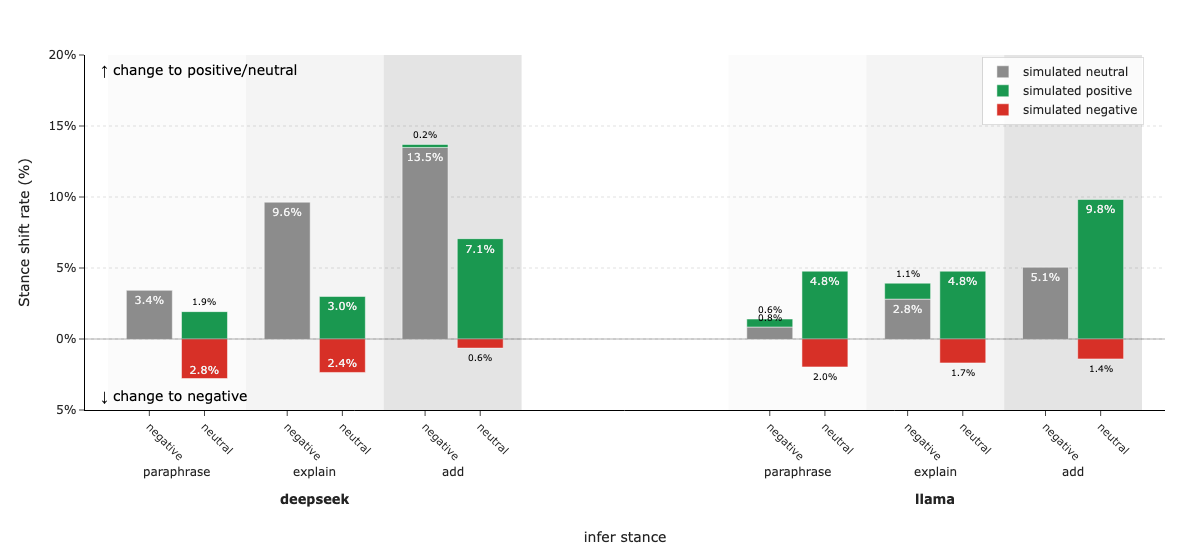}
    \caption{Stance shift rates across Claude revision strategies}
    \label{fig:claude-transition}
\end{figure*}

The corresponding results using Claude-based revisions and GPT-5.2 stancing are presented in Figure~\ref{fig:claude-transition}. Similar trends are observed across all text-only revision strategies.

\section{Sensitivity Analysis}\label{app:sensitive}
\subsection{Prompt and Temperature Sensitivity Analysis}
\label{subapp:prompt-temp-sens}

To evaluate the robustness of our stance simulation framework during text-only strategy revisions, we specifically examine whether changes in prompting style or decoding temperature substantially affect the resulting stance interpretations. For each revision strategy, we generate revision conversations under several configurations, including temperature 0.5, temperature 1.0, and two alternative paraphrased prompts. We use the default configuration (temperature 0 with the original prompt) as the reference setting in calculating accuracy. As shown in Tables~\ref{tab:combined_accuracy_gemini} and \ref{tab:combined_accuracy_claude}, high consistencies are observed across both temperatures and prompts.

\begin{table}[h]
\centering
\small
\resizebox{\columnwidth}{!}{
\begin{tabular}{lcccc}
\toprule
\textbf{Strategy} & \textbf{Temp 0.5} & \textbf{Temp 1} & \textbf{Prompt 2} & \textbf{Prompt 3} \\
\midrule
Add        & 87.42 & 86.33 & 87.15 & 87.42 \\
Explain    & 88.69 & 88.25 & 88.47 & 89.18 \\
Paraphrase & 92.97 & 91.76 & 92.31 & 92.92 \\
\bottomrule
\end{tabular}}
\caption{Combined stance agreement accuracy (\%) under Gemini revision across all three source models.}
\label{tab:combined_accuracy_gemini}
\end{table}

\begin{table}[h]
\centering
\small
\resizebox{\columnwidth}{!}{
\begin{tabular}{lcccc}
\toprule
\textbf{Strategy} & \textbf{Temp 0.5} & \textbf{Temp 1} & \textbf{Prompt 2} & \textbf{Prompt 3} \\
\midrule
Add        & 90.72 & 89.71 & 90.02 & 90.57 \\
Explain    & 91.27 & 90.41 & 90.88 & 90.41 \\
Paraphrase & 92.28 & 92.21 & 92.13 & 92.44 \\
\bottomrule
\end{tabular}}
\caption{Combined stance agreement accuracy (\%) under Claude revision across Llama and DeepSeek.}
\label{tab:combined_accuracy_claude}
\end{table}

\begin{table}[h]
\centering
\resizebox{\columnwidth}{!}{
\begin{tabular}{@{}llccc@{}}
\toprule
\textbf{Topic} & \textbf{\begin{tabular}[c]{@{}l@{}}Revision\\ Type\end{tabular}} & \textbf{\begin{tabular}[c]{@{}c@{}}Low-to-High\\ Shift (\%)\end{tabular}} & \textbf{\begin{tabular}[c]{@{}c@{}}High-to-Low\\ Shift (\%)\end{tabular}} & \textbf{\begin{tabular}[c]{@{}c@{}}Overall Depolarize\\ Rate (\%)\end{tabular}} \\ \midrule
Nuclear energy & Add & 87.2 & 84.5 & 86.3 \\
Nuclear energy & Meme & 75.2 & 36.5 & 62.1 \\ \midrule
Keto diet & Add & 87.2 & 84.5 & 86.2 \\
Keto diet & Meme & 24.7 & 82.8 & 47.0 \\ \midrule
Universal basic income & Add & 68.6 & 89.5 & 76.8 \\
Universal basic income & Meme & 46.9 & 72.9 & 56.6 \\ \midrule
SpaceX IPO & Add & 79.2 & 94.8 & 84.9 \\
SpaceX IPO & Meme & 33.3 & 75.0 & 50.5 \\ \bottomrule
\end{tabular}}
\caption{Tone-shift depolarization rates of the \textit{add} and \textit{meme} revisions on the four additional domains.}
\label{tab:app-domains-tone}
\end{table}

\subsection{Meme Template Sensitivity Analysis}\label{subapp:meme-sens}
To evaluate the robustness of our stance simulation framework during multimodal strategy revisions, we analyse the average directional stance shift across 5 meme templates. As shown in Table~\ref{tab:app-meme-sens}, high consistency is observed across meme templates within the topic.

\begin{table*}[t]
\centering
\begin{tabular}{@{}lcccccc@{}}
\toprule
\textbf{Topic} & \textbf{Inferred} & \textbf{meme\_1 \(\Delta\)} & \textbf{meme\_2 \(\Delta\)} & \textbf{meme\_3 \(\Delta\)} & \textbf{meme\_4 \(\Delta\)} & \textbf{meme\_5 \(\Delta\)} \\ \midrule
Claude & 0.2358 & +0.1358 & +0.1547 & +0.1283 & +0.1472 & +0.1302 \\
Deepseek & 0.1705 & +0.0891 & +0.0649 & +0.0509 & +0.0840 & +0.0891 \\
Llama & 0.1301 & +0.0650 & +0.0569 & +0.0407 & +0.0691 & +0.0285 \\ \bottomrule
\end{tabular}
\caption{The average directional stance shift of different meme templates on all target topics.}
\label{tab:app-meme-sens}
\end{table*}

\section{Generalization to Additional Domains}\label{app:domains}

To test whether the revision effects generalize beyond AI-related discussions, we extend the experiments to four new domains with four new stance targets: \textbf{nuclear energy} (climate/energy), \textbf{the keto diet} (health/lifestyle), \textbf{universal basic income} (politics/economics), and \textbf{the SpaceX IPO} (finance). For each target we collect 300 Reddit conversations (1,200 in total) using the same collection protocol as in Appendix~\ref{appendix_sec:data}, and run the identical pipeline, prompt templates, revision models, and stance simulator as in the main experiments.

\paragraph{Revision effectiveness.} Table~\ref{tab:app-domains-delta} reports the average directional stance shift on the four new domains, using the same metric as Table~\ref{tab:main-delta}. The cross-domain averages closely track the main results (\textit{add}: +49.8\% vs. +44.8\%; \textit{meme}: +41.3\% vs. +49.3\%; \textit{explain}: +25.2\% vs. +17.5\%; \textit{paraphrase}: -2.5\% vs. -4.0\%), with the same strategy ordering (\textit{add} $\approx$ \textit{meme} $>$ \textit{explain} $\gg$ \textit{paraphrase} $\approx$ 0) and all effective strategies significant at $p<0.001$ on average. The consistent replication across four additional domains demonstrates that the effects revealed by our counterfactual auditing framework are general rather than specific to AI-related discussions.

\paragraph{Depolarization mechanism.} Table~\ref{tab:app-domains-tone} recomputes the LIWC Tone-shift analysis with the same definitions as Table~\ref{tab:meme-antipolar}. The pattern matches the main results: \textit{add} consistently depolarizes tone in both directions (overall 76.8--86.3\%, vs. 78.7--81.7\% in the main experiments), whereas \textit{meme} shifts tone close to randomly (47.0--62.1\% vs. 54.5--58.6\% in the main experiments). This consistency shows that the mechanism-level account in Section~\ref{sec:results_mechanism} holds across domains rather than being an artifact of AI-related topics.

\begin{table*}[t]
\centering
\small
\setlength{\tabcolsep}{4pt}
\begin{tabular}{@{}lccccc@{}}
\toprule
\textbf{Target Topic} & \textbf{Inferred} & \textbf{paraphrase \(\Delta\)} & \textbf{explain \(\Delta\)} & \textbf{add \(\Delta\)} & \textbf{meme \(\Delta\)} \\ \midrule
Nuclear energy & +0.077 & -0.003 (-4.3\%) & +0.033 (+43.5\%)$^{*}$ & +0.063 (+82.6\%)$^{***}$ & +0.054 (+70.5\%)$^{***}$ \\
Keto diet & +0.177 & -0.007 (-3.8\%) & +0.027 (+15.1\%) & +0.040 (+22.6\%)$^{*}$ & +0.057 (+32.1\%)$^{***}$ \\
Universal basic income & +0.153 & -0.007 (-4.3\%) & +0.030 (+19.6\%)$^{*}$ & +0.047 (+30.4\%)$^{***}$ & +0.041 (+26.4\%)$^{*}$ \\
SpaceX IPO & -0.147 & +0.003 (+2.3\%) & +0.033 (+22.7\%) & +0.093 (+63.6\%)$^{***}$ & +0.053 (+36.4\%)$^{*}$ \\ \midrule
Average & -- & -0.003 (-2.5\%) & +0.031 (+25.2\%)$^{***}$ & +0.061 (+49.8\%)$^{***}$ & +0.051 (+41.3\%)$^{***}$ \\ \bottomrule
\end{tabular}
\caption{The average directional stance shift of different strategies on the four additional domains, with the relative change over the inferred baseline in parentheses. Significance markers denote a paired sign test of each strategy's per-conversation stance shift against the \textit{paraphrase} control on the same conversations ($^{*}p<0.05$, $^{**}p<0.01$, $^{***}p<0.001$); the Average row pools all 1,200 conversations.}
\label{tab:app-domains-delta}
\end{table*}

\section{Meme Templates}\label{app:meme-temp}
\begin{figure*}[h]
    \centering

    \begin{subfigure}[t]{0.3\textwidth}
        \centering
        \includegraphics[width=\linewidth]{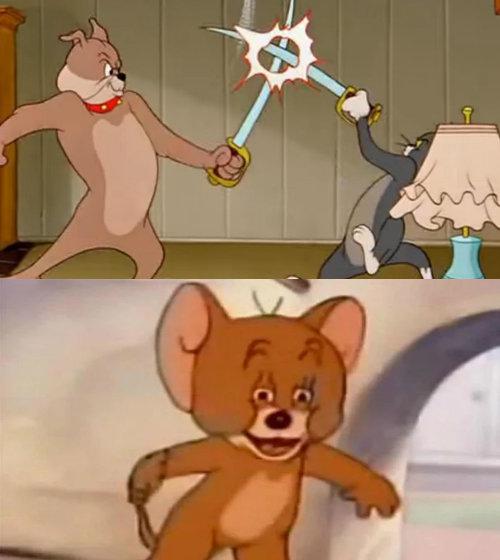}
        \caption{Meme template 1}
        \label{subfig:meme-1}
    \end{subfigure}
    \hfill
    \begin{subfigure}[t]{0.3\textwidth}
        \centering
        \includegraphics[width=\linewidth]{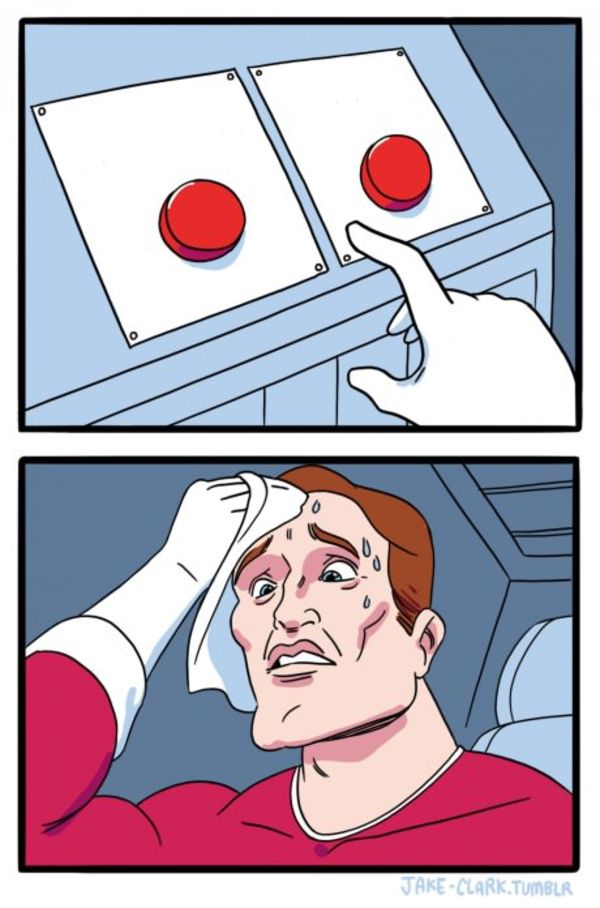}
        \caption{Meme template 2}
        \label{subfig:meme-2}
    \end{subfigure}
    \hfill
    \begin{subfigure}[t]{0.3\textwidth}
        \centering
        \includegraphics[width=\linewidth]{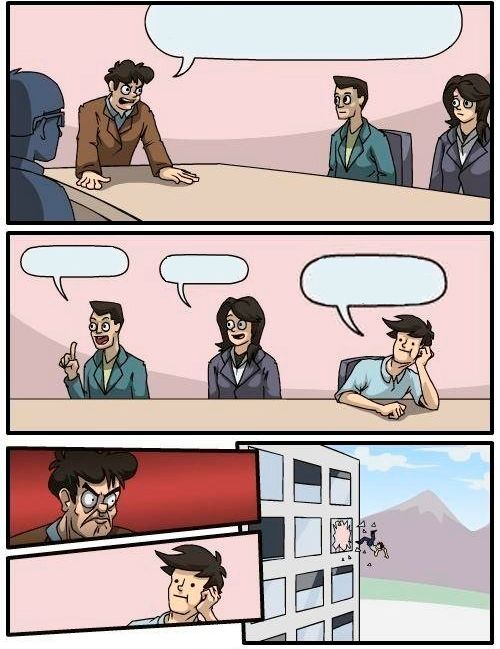}
        \caption{Meme template 3}
        \label{subfig:meme-3}
    \end{subfigure}
    
    \vspace{0.5em}

    \begin{subfigure}[t]{0.3\textwidth}
        \centering
        \includegraphics[width=\linewidth]{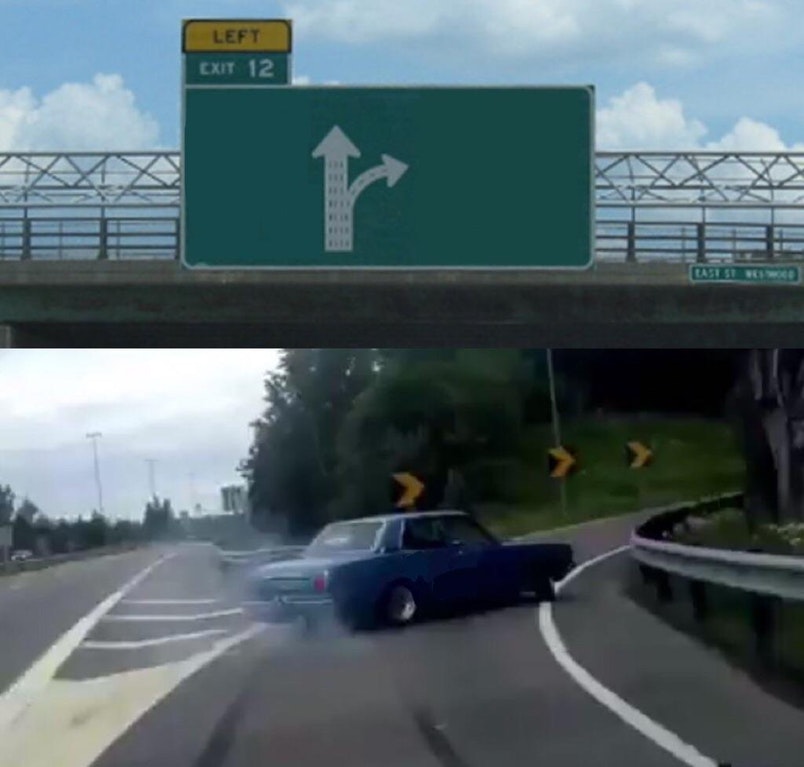}
        \caption{Meme template 4}
        \label{subfig:meme-4}
    \end{subfigure}
    \hspace{0.03\textwidth}
    \begin{subfigure}[t]{0.3\textwidth}
        \centering
        \includegraphics[width=\linewidth]{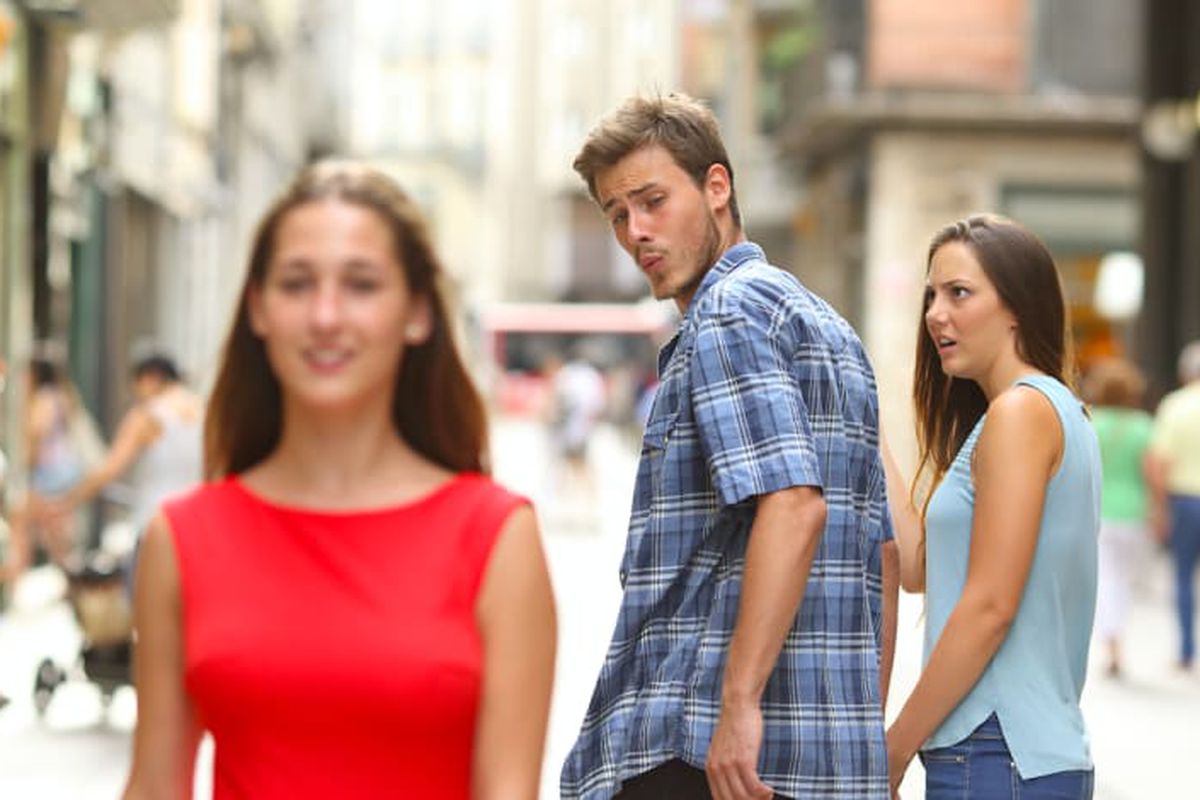}
        \caption{Meme template 5}
        \label{subfig:meme-5}
    \end{subfigure}
    
    \caption{Meme templates used in the counterfactual context revision.}
    \label{fig:meme-temp-all}
\end{figure*}

We provide all the meme templates in Figure~\ref{fig:meme-temp-all}, collected from top-ranking templates on ImgFlip.

\FloatBarrier
\raggedbottom

\section{Revision Prompts}
\label{app:revision-prompt}

\subsection{Paraphrase Prompt}
\begin{tcolorbox}[width=\columnwidth, breakable]
\begin{lstlisting}[caption=Paraphrase Revision Strategy]
Paraphrase the following message while preserving its original meaning and tone. Only output the revised message. Do not include anything else.
\end{lstlisting}
\end{tcolorbox}

\subsection{Explain Prompt}

\begin{tcolorbox}[width=\columnwidth, breakable]
\begin{lstlisting}[caption=Explain Revision Strategy]
You will be given a multi-party conversation about {target_model}. The last message is from [OTHER USER]. Your goal is to revise ONLY that last message so that it clearly responds to, explains, and addresses the concerns raised by the TARGET USER.

Target user: [TARGET USER]
Last message speaker: [OTHER USER]

Your task:
- Identify the concerns raised by the target user in the conversation.
- Revise ONLY the last message in a way that:
   - directly addresses those concerns
   - provides clear explanations
   - acknowledges misunderstandings or ambiguities
   - maintains factual accuracy
- It should avoid false claims or fabricated information.
- It should avoid being manipulative or overtly persuasive.
- It should use acknowledgment, clarification, balanced reasoning, and constructive framing to guide the target user gently toward a more positive or neutral stance.

Only output the revised message. Do not include anything else.

<conversation transcript below>
{conv_transcript}
<last message below>
{last_message}
\end{lstlisting}
\end{tcolorbox}

\subsection{Add Prompt}

\begin{tcolorbox}[width=\columnwidth, breakable]
\begin{lstlisting}[caption=Add Revision Strategy]
You will be given a multi-party conversation about {target_model}. The last message is from [OTHER USER]. Your goal is to revise ONLY that last message so that it includes additional arguments designed to persuade the TARGET USER toward a more positive or neutral stance on {target_model}.

Target user: [TARGET USER]
Last message speaker: [OTHER USER]

Your task:
1. Analyze the concerns, objections, or negative signals expressed by the target user.
2. Identify opportunities where additional arguments could constructively address or counter those concerns.
3. Revise ONLY the last message so that it:
   - adds **new but reasonable arguments** not previously mentioned in the conversation
   - remains factually accurate (no false claims)
   - stays consistent with the conversational context
   - responds directly to the target user's concerns
   - uses a tone that is respectful, clear, and non-manipulative
   - is aimed at shifting the target user's stance toward positive or neutral
4. You may introduce new reasoning or perspectives, but you must NOT introduce unverifiable facts.
5. Do NOT modify earlier messages.
6. Do NOT contradict anything stated earlier in the conversation.

Only output the revised message. Do not include anything else.

<conversation transcript below>
{conv_transcript}
<last message below>
{last_message}
\end{lstlisting}
\end{tcolorbox}

\subsection{Meme Prompt}
\begin{tcolorbox}[width=\columnwidth, breakable]
\begin{lstlisting}[caption=Meme Text Generation]
You will be given a multi-party conversation about target_model} and a meme template. Your task is to reply to the TARGET USER with an argument that change his stance, and embed it into the meme template to create a new meme.

Instructions:
1. Read the conversation to understand the context and the concerns of the TARGET USER.
2. Act as if you are the last [OTHER USER] and come up with a reply to change the TARGET USER's stance to positive.
3. Adapt that reply into concise, punchy meme text format that fits the structure and humor style of the given template.
4. Based on the meme template's structure, determine the appropriate text positions (e.g. "top_text", "bottom_text", "panel_1", "panel_2", "caption", "left", "right", etc.) and output a JSON object where each key represents a text position in the template and each value is the corresponding meme text.
   - Use position names that naturally reflect the template's layout and format.
   - Only include positions that the template actually has.
   - Keep each text segment short (ideally under 10 words) and impactful.
5. Do not include any explanation or extra output, only the JSON.

<conversation transcript below>
{conv_transcript}
\end{lstlisting}
\end{tcolorbox}

\begin{tcolorbox}[width=\columnwidth, breakable]
\begin{lstlisting}[caption= Meme Generation]
Add the text to the meme template according to the following instructions:
f"{meme_context}
Only add the text at the specified location according to the instructions; DO NOT change anything else in the image. This is only for research purposes.
\end{lstlisting}
\end{tcolorbox}

\section{Stance Prompts}\label{app:stance-prompt}

\begin{tcolorbox}[width=\columnwidth, breakable]
\begin{lstlisting}[caption= Observed Stance]
Instruction: 
You will be given a multi-party conversation that includes messages from a target user and other users. Your task is to determine the target user's stance toward {target_model}.

Input:
* A conversation transcript involving multiple users
* The target user is explicitly identified as: [TARGET USER]
* Other users may express opinions, but your output must be based only on the target user's own statements, not others.

Your task:
* Read the conversation.
* Focus exclusively on the statements made by the target user.
* Infer the target user's stance toward {target_model}.
* Base your inference on explicit or strongly implied sentiment from the target user.

Output Format:
* Provide your answer the following JSON format: 
{{"stance": positive | neutral | negative, "reasoning": reason}}

<conversation transcript below>
{conv_transcript}
\end{lstlisting}
\end{tcolorbox}

\begin{tcolorbox}[width=\columnwidth, breakable]
\begin{lstlisting}[caption= Inferred Stance]
Instruction: 
You will be given a multi-party conversation that includes messages from a target user and other users. Your task is to PREDICT the target user's stance toward {target_model} AFTER having this conversation.

Input:
* A conversation transcript involving multiple users
* The target user is explicitly identified as: [TARGET USER]

Your task:
* Read the conversation.
* Predict the target user's stance (positive, neutral, or negative) toward {target_model} AFTER the conversation.
* Consider both:
  (a) what the target user explicitly said, and
  (b) how the discussion context, arguments, tone, and interactions might influence the target user's stance.
* If there is insufficient evidence to determine a change in stance, classify the stance as neutral.
* Do not treat other users' stances as the target user's stance; only use them to infer potential influence.
* Make a prediction, not a summary.

Output Format:
* Provide your answer the following JSON format: 
{{"stance": positive | neutral | negative, "reasoning": reason}}

<conversation transcript below>
{conv_transcript}
\end{lstlisting}
\end{tcolorbox}

\end{document}